\documentclass[aoas,MSNbibl,nameyear,seceqn,dvips]{arximspdf}
\usepackage{mathbh}
\usepackage{graphicx}

\doi{10.1214/13-AOAS698} 
\volume{8}
\issue{1}
\pubyear{2014}
\firstpage{499}
\lastpage{529}

\makeatletter
\newcommand{\rrVert}{\Vert}
\newcommand{\rrvert}{\vert}
\newcommand{\llVert}{\Vert}
\newcommand{\llvert}{\vert}
\def\mathds{\mathbh}
\makeatother

\begin{document}
\begin{frontmatter}

\title{Concise comparative summaries (CCS) of large text corpora with a
human experiment\thanksref{T1}}
\runtitle{Concise comparative summaries (CCS)}

\begin{aug}
\author[a]{\fnms{Jinzhu} \snm{Jia}\corref{}\thanksref{m2,t2}\ead[label=e2]{jzjia@math.pku.edu.cn}},
\author[b]{\fnms{Luke} \snm{Miratrix}\thanksref{m1,t2}\ead[label=e1]{lmiratrix@stat.harvard.edu}},
\author[c]{\fnms{Bin} \snm{Yu}\thanksref{m3}\ead[label=e3]{binyu@stat.Berkeley.EDU}},
\author[d]{\fnms{Brian} \snm{Gawalt}\thanksref{m3}\ead[label=e4]{bgawalt@gmail.com}},
\author[d]{\fnms{Laurent} \snm{El Ghaoui}\thanksref{m3}\ead[label=e5]{elghaoui@berkeley.edu}},
\author[e]{\fnms{Luke} \snm{Barnesmoore}\thanksref{m4}\ead[label=e6]{lrb@mail.sfsu.edu}}
\and
\author[e]{\fnms{Sophie} \snm{Clavier}\thanksref{m4}\ead[label=e7]{sclavier@sfsu.edu}}

\runauthor{J. Jia et al.}

\thankstext{T1}{Supported in part by NSF Grant SES-0835531 under the
``Cyber-Enabled Discovery and Innovation (CDI),'' NSF Grant DMS-09-07632,
ARO Grant W911NF-11-1-0114, NSF Grant CCF-0939370, NSF-CMMI Grant 30148,
NSFC-11101005 and DPHEC-20110001120113.}

\thankstext{t2}{J. Jia and L. Miratrix are co-first authors and are listed in alphabetical order.}

\affiliation{Peking University\thanksmark{m2},
Harvard University\thanksmark{m1},
University of California, Berkeley\thanksmark{m3}\break
and San Francisco State University\thanksmark{m4}}

\address[a]{J. Jia\\
LMAM, School of Mathematical Sciences\\
\quad and Center for Statistical Science\\
Peking University\\
Beijing 100871\\
China\\
\printead{e2}}

\address[b]{L. Miratrix\\
Department of Statistics\\
Harvard University\\
1 Oxford Street\\
Cambridge, Massachusetts 02138-2901\\
USA\\
\printead{e1}}

\address[c]{B. Yu\\
Department of Statistics\\
\quad and Department of EECS\\
University of California, Berkeley\\
Berkeley, California 94720\\
USA\\
\printead{e3}}

\address[d]{B. Gawalt\\
L. El Ghaoui\\
Department of EECS\\
University of California, Berkeley\\
Berkeley, California 94720\hspace*{44pt}\\
USA\\
\printead{e4}\\
\phantom{E-mail:\ }\printead*{e5}}

\address[e]{L. Barnesmoore\\
S. Clavier\\
Department of International Relations\\
College of Liberal \& Creative Arts\\
San Francisco State University\\
San Francisco, California 94132\\
USA\\
\printead{e6}\\
\phantom{E-mail:\ }\printead*{e7}}
\end{aug}

\received{\smonth{2} \syear{2013}}
\revised{\smonth{10} \syear{2013}}

%
\begin{abstract}
In this paper we propose a general framework for topic-specific
summarization of large text corpora and illustrate how it can be used
for the analysis of news databases.
Our framework, concise comparative summarization (CCS), is built on
sparse classification methods.
CCS is a lightweight and flexible tool that offers a compromise between
simple word frequency based methods currently in wide use and more
heavyweight, model-intensive methods such as latent Dirichlet
allocation (LDA).
We argue that sparse methods have much to offer for text analysis and
hope CCS opens the door for a new branch of research in this important field.

For a particular topic of interest (e.g., China or energy), CSS
automatically labels documents as being either on- or off-topic
(usually via keyword search), and then uses sparse classification
methods to predict these labels with the high-dimensional counts of all
the other words and phrases in the documents.
The resulting small set of phrases found as predictive are then
harvested as the summary.

To validate our tool, we, using news articles from the New York Times
international section, designed and conducted a human survey to compare
the different summarizers with human understanding.
We demonstrate our approach with two case studies, a media analysis of
the framing of ``Egypt'' in the New York Times throughout the Arab
Spring and an informal comparison of the New York Times' and Wall
Street Journal's coverage of ``energy.''
Overall, we find that the Lasso with $L^2$ normalization can be
effectively and usefully used to summarize large corpora, regardless of
document size.
\end{abstract}

%

\end{frontmatter}

\section{Introduction}
\label{Sec:intro}

Stuart Hall\setcounter{footnote}{2}\footnote{Quoted in James Watson's 2007 article,
``Representing Realities: An Overview of News Framing.''}
wrote, ``the media are part of the dominant means of ideological
production. What they produce is precisely representations of the
social world, images, descriptions, explanations and frames for
understanding how the world is and why it works as it is said and shown
to work.'' Given this, in order to understand how the public constructs
its view of the world, we need to be\vadjust{\goodbreak} able to generate concise,
comprehensible summaries of these representations.
Automatic, concise summaries thus become quite useful for comparing
themes across corpora or screening corpora for further readings.

Our approach to obtain such summaries is by first identifying a corpus
that we believe contains substantial information on prespecified topics
of interest and then using automated methods to extract summaries of
those topics. These summaries ideally show the connections between our
topics and other concepts and ideas.
The two corpora we investigate in this paper are all the articles in
the international section of the New York Times from 2009 to just after
2011, and all the headlines from both the New York Times and the Wall
Street Journal from 2008 to 2011.
Our approach, however, could be applied to other corpora, such as the
writings of Shakespeare, books published in statistics in 2012
or Facebook wall writings of some community.
Since such corpora are large, only a very tiny fraction of them could
ever be summarized or read by humans.

There are many ways one might study a corpus.
One common and effective method for text study is comparison.
For example, a media analyst interested in investigating how the topic
of ``China'' is framed or covered
by NYT's international section in 2009 could form an opinion by
comparing articles about China to those not about China.
A Shakespeare scholar could gain understanding on Shakespeare's view on romance
by comparing the author's romantic plays with his nonromantic plays.

In this paper, we propose and validate by human survey a topic-driven concise
comparative summarization (CCS) tool for large text corpora.
Our CCS tool executes the comparison idea through statistical
sparse ``classification'' methods.
We first \emph{automatically} label blocks of text in a corpus as
``positive'' examples about a topic or ``negative'' (``control'') examples.
We then use a machine learning predictive framework and sparse
regression methods such as the Lasso [\citet{tibshirani1996regression}]
to form a concise summary of the positive examples out of those phrases
selected as being predictive of this labeling.

A novel advantage of our tool is the flexible nature of its labeling process.
It allows different ways of forming ``positive'' and ``negative''
examples to provide
``snapshot'' summaries of a corpus from various angles.
For instance, we could label articles that mention China as
``positive'' examples and the rest as
``negative examples;'' we could also take the same positive examples and
use only those articles that contain other Asian countries (but not
China) as the negative examples.
Because the summaries are concise, it is possible for researchers to
quickly and effectively examine and compare multiple snapshots.
Therefore, changes in coverage across time or between sources can be
presented and understood even when the changes are multidimensional and complex.

Even though our tool takes a classification framework as its
foundation, our interest is in understanding text rather than
classifying it.
Therefore, we validated our tool through a systematic randomized human
survey, described in Section~\ref{sec:humanexperiment}, where human
subjects evaluated our summaries based on their reading of samples from
the original text.
This provided some best practices for generating summaries with the
highest overall quality as measured by, essentially, relevance and clarity.

Our CCS tool can be used to provide confirmatory evidence to support
pre-existing theories.
Extending the work of \citet{chinapaper}, in Section~\ref{sec:casestudy}
media analyst co-authors of this paper use this tool and framing theory
(an analytical framework from media studies, described later) to
compare the evolution of news media representations of countries across
different distinct periods defined by significant events such as
revolutionary upheaval or elections with existing international
relations theory.
Our tool can also be used to explore text in a generative manner,
helping researchers better understand and theorize about possible
representations or framing mechanisms of a topic in a body of text.
In our second case study we utilize CCS to compare the headlines of the
New York Times to the Wall Street Journal, in particular, for the topic
of ``energy.''

The rest of the paper is organized as follows.
Before presenting our proposed approach, concise comparative summary
(CCS), we
briefly review related work in Section~\ref{Sec:related works}.
Section~\ref{Sec:CCS} describes the CCS framework, which consists of
three steps:
\begin{longlist}[1.]
\item[1.] the labeling scheme: what rule to use
to \emph{automatically} label a document unit as ``positive'' or ``negative;''
\item[2.] preprocessing: when building and expanding on a \emph{bag of
words} representation of a corpus, we must decide
which document unit to use (article vs. paragraph) and
how to rescale counts of phrases appropriately; and
\item[3.] feature selection:
how to select the summary phrases.
\end{longlist}
For preprocessing, we describe tf-idf and $L^2$ rescaling. For feature
selection, we discuss the Lasso, $L^1$-penalized logistic regression
(L1LR), correlation and co-occurrence.
Note that the former two fall into the predictive framework, while the
last do not but
are included because of their wide use.
The human validation experiment to compare different combinations in
the CCS framework over labeling, rescaling, unit choice and feature
selection choice is described in Section~\ref{sec:humanexperiment} with
results in Section~\ref{sec:humanvalidresults}.
Section~\ref{sec:casestudy} presents the two case studies introduced
above, using the Lasso with $L^2$ normalization, the method found to be
the most robust in the human validation experiment.
Section~\ref{sec:conclusion} concludes with a discussion.

\section{Related works}
\label{Sec:related works}

Automated tools aimed at understanding text, especially newspaper text,
are becoming more and more important with the increased accumulation
of text documents in all fields of human activities.
In the last decade we have seen the emergence of computational social
science, a field
connecting statistics and machine learning to anthropology, sociology,
public policy and
more [\citet{lazeretal09}]. Automatic summarization is in wide use:
Google news trends, Twitter's trending topics [\citet{Zubiaga11}] and
Crimson Hexagon's brand analysis all use text summaries to attempt to
make sense of the vast volumes of text generated in public discourse.
These all illustrate the great potential of statistical methods for
text analysis, including news media analysis.
We hope our proposed CCS framework will help advance this new and
exciting field.

Most text summarization approaches to date (aside from natural
language- and grammar-based approaches) use word or phrase
(including sentence) counts or frequencies. They can be
considered along two axes. The first axis is whether an approach
generates topics on its own or summarizes without regard to topic
(unsupervised) or is supplied a topic of interest (supervised).
The second axis is whether the word and phrase rates of appearance are
modeled or simply reweighted.

\subsection{Unsupervised model-based approaches}
\label{Sec:topic}

Topic modeling, where documents in a corpus are described as mixtures
of latent topics that are in turn described by words and phrases, is a
rapidly growing area of text analysis. These methods take text
information as input and produce a
(usually generative) model fit to the data. The model itself captures
structure in the data, and this structure can be viewed as a summary.
The set of topics generated can serve as a summary of the corpus
overall, and individual documents can be summarized by presenting those
topics most associated with them.

A popular example is the latent Dirichlet allocation (LDA) model [\citet
{Blei03}], which posits that each word observed in the text stands in
for a hidden, latent ``topic'' variable.
These models are complex and dense: all words play a role in all the
topics. However, one can still present the most prominent words in a
topic as the summary, which produces cogent and reasonable topics; see
\citet{chang09}, where humans evaluate the internal cohesion of learned
topics by identifying ``impostor'' words inserted into such lists.
\citet{grimmeretal2011} combine such a model with clustering to organize
documents by their topics. They also extensively evaluate different
models under their framework with human survey experiments.

Summarizing or presenting the generated topics with this method can be
problematic. For example, taking the most probable words of a topic to
represent it can lead to overly general representations. \citet
{bischof2012} propose focusing on how words discriminate between topics
as well as overall frequency---essentially a comparative approach---to
better identify overall topics. These issues notwithstanding, LDA-style
approaches are quite powerful and can be used comparatively. For
example, \citet{paul2010summarizing} use LDA to score sentences from
opposite viewpoints to summarize differences between two ideologies.

\subsection{Unsupervised simple weighting approaches}
\label{Sec:simple}

Google Trends charts are calculated by comparing the number of times a
prespecified
word of interest appears to the overall volume of news for a specified
time period (within the news outlets that Google compiles).
Even this simple approach can show how topics enter and leave public
discourse across time. Twitter's trending topics appear to operate
similarly, although it selects the hottest topics by those which are
gaining in frequency most quickly.
These approaches are similar in spirit to the normalized simpler
methods (co-occur and correlation screen) that we compare with CCS in
this paper.

\citet{HopKin10} extrapolate from a potentially nonrandom sample of
hand-coded documents to estimate the proportion of documents in several
predefined categories. This can
be used for sentiment analysis (e.g., estimating the proportion of
blogs showing approval for some specified public figure). Their work
drives Crimson Hexagon, a company currently offering brand analysis to
several companies. Our approach instead identifies key phrases most
associated with a given topic or subject.

There is a wide literature on text summarization (as compared to topic
modeling, above) by key-phrase extraction [\citet
{roseetal2010,senellart2008,franketal1999}] and sentence
extraction \citet{hennig2009,goldstein2000,netoetal2002}.
These approaches score
potential key phrases or sentences using metrics such as position in a
paragraph, sentence length or frequency of occurrence, and then select the
highest scorers as the summary.
While typically used for individual documents, \citet{goldstein2000}
did extend this approach to multiple documents by scoring and selecting
sentences sequentially, with future sentences penalized by similarity
to previously selected sentences.

In \citet{monroe08}, the authors take a comparative approach as we do.
They merge all text into two super-documents (the positive and negative
examples) and then score individual words based on their rates of
appearance normalized by their overall frequency. We analyze the corpus
through individual document units.

\subsection{Supervised approaches}
\label{Sec:supervised}

Supervised versions of LDA that incorporate a given topic labeling in
the hierarchical Bayesian model [\citet{BleiMcA08}] do exist.
Although these methods are computationally expensive and produce dense
models requiring truncation for interpretability, they are powerful
indications of the capabilities of computer-assisted topic-based
summarization. \citet{hennig2009} applies a latent topic model similar
to LDA for topic-specific summarization of documents. Here the topic is
represented as a set of documents and a short narrative of the desired
content and sentences are then extracted by a scoring procedure that
compares the similarity of latent sentence representations to the
provided topic of interest.

\emph{Classification} of text documents using the phrases in those
documents as features (and a given, prespecified labeling of those
documents) is familiar and well studied [\citet{genkin07,zhang01}].
However, while we extensively build on this work, our focus is \emph
{not} on the ability to classify documents but rather on the
interpretable features that enable classification.
Interpreting these features allows for investigation of the quality of
the text in relation to other variables of interest. For example, \citet
{eisenstein2011discovering} use similar approaches to examine the
relationship between characteristics of different authors and their
patterns of lexical frequencies.

\section{Our approach: Concise comparative summarization (CCS) via
sparse predictive classification}
\label{Sec:CCS}

In science and engineering applications, statistical models often lend
themselves to believable generative stories.
For social science applications such as text analysis, however, models
are more likely to be descriptive than generative.
As simple methods are more transparent, they are arguably more
appealing for such descriptive purposes.
Our overall goal is to develop computationally light as well as
transparent tools for text analysis and, by doing so, to explore the
limits of methods that are not extensively model-based.

\begin{table}[b]
\tabcolsep=4pt
\caption{Four different countries in 2009. The method used (a count
rule with a threshold of 2, the Lasso for feature selection, and tf-idf
reweighting of features) was one of the best identified for
article-unit analysis by our validation experiment}
\label{tab:foursummaries}
\begin{tabular*}{\textwidth}{@{\extracolsep{\fill}}lccc@{}}
\hline
\textbf{Iraq} & \textbf{Russia} & \textbf{Germany} & \textbf{Mexico} \\
\hline
american & a medvedev & angela merkel & and border protection \\
and afghanistan & caucasus & berlin & antonio betancourt \\
baghdad & europe & chancellor angela & cancn \\
brigade & gas & european & chihuahua \\
combat & georgia & france and & denise grady \\
gen & interfax news agency &
frankfurt & drug cartels \\
in afghanistan & iran & group of mostly & guadalajara \\
invasion & moscow & hamburg & influenza \\
nuri & nuclear & marwa alsherbini & oaxaca \\
pentagon & president dmitri & matchfixing & outbreak \\
saddam & republics & minister karltheodor zu & president felipe \\
sergeant & sergei & munich & sinaloa \\
sunni & soviet & nazi & swine \\
troops & vladimir & world war & texas \\
war and who & & & tijuana \\
\hline
\end{tabular*}
\end{table}

Our CCS framework is composed of three main steps:
\begin{longlist}[1.]

\item[1.] automatically label the text units for a given topic (label),

\item[2.] preprocess the possible summarizing phrases and phrase counts
(weight), and

\item[3.] sparsely select a comparative phrase
list of interest using classification methods on the automatic labels
(summarize).
\end{longlist}

For a given topic or subject (e.g., ``Egypt'') in a given context (e.g.,
the NYT international section in 2009), CCS produces summaries in the
form of a list of key phrases.
To illustrate, Table~\ref{tab:foursummaries} contains four sample summaries.
Here we labeled an article as a ``positive'' example if it contains the word
of the country under various forms at least twice.
As we can see in this table,
sometimes fragments are selected as stand-ins for complete phrases, for
example, the phrase ``president felipe'' appears in the Mexico column,
signifying \emph{President Felipe [Calder\`{o}n]}. These summaries are
suggestive of the aspects of these countries that
are most covered in the New York Times in 2009, relative
to other topics: even now, ``nazis'' and the ``world wars'' were tied
to Germany;
``iraq'' and ``afghanistan'' were also tied closely;
``gen'' (as in the military title \emph{General}) and ``combat'' were
the major focus in Iraq.
The coverage of Mexico revolved around the ``swine flu,'' ``drug
cartels'' and concerns about the ``border.'' Russia had a run-in with
Europe about ``gas,'' and ``nuclear'' involvement with ``iran.''

We use sparse classification tools such as the Lasso or $L^1$-penalized
logistic regression (L1LR) in step 3; these are fast and different from
the modeling methods described earlier.
Our approach is fundamentally about contrasting sets of documents and
using found differences as the relevant summary, which allows for a
more directed process of summarization than unsupervised methods.
This also allows for multiple snapshots of the same topic in the same
document corpus using different contrasting sets, which gives a more
nuanced understanding of how the topic is portrayed.

%

To situate concise comparative summarization of a given topic in a
binary classification framework, we now introduce some notation.
A predictive framework consists of $n$ units, each with a class label
$y_i \in\{-1, +1\}$ and a collection of $p$ possible features that can
be used to predict this class label. Each unit $i \in\mathcal{I}
\equiv\{ 1,\ldots,n \}$ is attributed a value $x_{ij}$ for each
feature $j \in\mathcal{J} \equiv\{ 1, \ldots, p \}$. These $x_{ij}$
form an $n \times p$ matrix $X$.
The $n$ units are blocks of text taken from the corpus (e.g., entire
articles or individual paragraphs), the class labels $y_i$ (generally
built automatically with keyword searches) indicate whether document
unit $i$ contains content on a subject of interest, and the features
are all the possible key phrases that could be used to summarize the
subject or topic.

$X$ is built from $C$, where $C$ is a representation of text often
called the \emph{bag-of-phrases model}: each document is represented as
a vector with the $j$th element being the total number of times that
the specific phrase $j$ appears in the document. Stack these row
vectors to make the \emph{document-term matrix} $C \in\mathbb{R}^{n
\times p}$ of counts.
From $C$, we build $X$ by rescaling the elements of $C$ to account for
different rates of appearance between the phrases.
$C$ and $X$ have one row for each document and one column for each
phrase, and they tend to be highly sparse: most matrix elements are 0.

Given the processed text $X$ and $y$, we can construct summarizers by
labeling, weighting and selecting phrases.
We can make different choices for each step.
We now present several such choices, and then discuss a human
validation experiment that identifies the best combination of these elements.

\subsection{Automatic and flexible labeling of text units}

To start, based on subject knowledge, the user of our tool (e.g., the
media analyst) translates a topic or subject of
interest into a set of topic phrases. For instance, he/she
might translate the topic of ``China'' into a topic list:
\emph{China, Chinas, Chinese}. Energy might be \emph{oil, gas,
electricity, coal, solar}. Arab Spring might be \emph{arab spring, arab
revolution, arab uprising}.\footnote{These topics can be refined and
expanded if initially generated summaries return other phrases that are
essentially the same. For example, in one of our case studies, we ran
CCS using the above \emph{energy} list as a query. When we saw the term
``natural'' surface as a summary word, we realized our query set could
be improved with the addition of the query \emph{natural gas}---CCS
helped us discover a useful addition to the query set, leading to a
broader, more useful summarization from a second pass using the
expanded query set. Topic modeling and keyword expansion methods could
also be of use here.}

Given a topic list, the user can apply different rules
to generate the labeling $y$. For example, label a text unit
as a ``positive,'' $+1$ example for the topic of ``China'' if the text unit
contains any of the phrases in the topic set, or, alternatively, if a
more stringent criterion is desired,
label it as ``positive'' if it contains more than two topic set phrases.

The general rules for labeling-by-query-count we used are as follows:
\begin{longlist}[\quad]
\item[\textit{count-K}]\hspace*{-5pt}: A document $i$ is given a label $y_i = +1$ if a query
term appears $K$ or more times in the document. Documents with $K-1$ or
fewer query hits receive a label of $y_i = -1$.
\item[\textit{hard-count-K}] or \emph{hcount-$K$}: As above, but drop all
documents with between 1 and $K-1$ hits from the analysis, as their
relationship to the query may be ambiguous.
\end{longlist}

In other cases labeling is straightforward. For directly comparing the
NYT to the WSJ, the labeling was $+1$ for NYT headlines and $-1$ for WSJ
headlines. For comparing a period of time to the rest, labeling would
be built from the dates of publication.

The labeling step identifies a set of documents to be summarized \emph
{in the context of} another set.
Generally, we summarize compared to the overall background of all
remaining documents, but one could drop ``uncertain'' documents, for
example, those with only one topic phrase but not more than one, or
``irrelevant'' ones, for example, those not relating to any Asian
country at all.
Different choices here can unveil different aspects of the corpus; see
Section~\ref{sec:nyt_wsj} for a case study that illustrates this.

\subsection{Preprocessing: Weighting and stop-word removal}
\label{sec:rescaling}

It is well known that baseline word frequencies impact information
retrieval methods and so raw counts are often adjusted to account for
commonality and rarity of terms [e.g., \citet{monroe08,salton88}].
In the predictive framework, this adjustment is done with the
construction of the feature matrix $X$.
We consider three constructions of $X$, all built on the bag-of-phrases
representation $C$.
Regardless of the weighting approach, we also remove any columns
corresponding to any phrases used to generate the labeling to prevent
the summary from being trivial and circular.
\citet{salton88} examine a variety of weighting approaches for document
retrieval in a multi-factor experiment and found choice of approach to
be quite important; we compare the efficacy of different choices in our
human validation survey (see Section~\ref{sec:humanexperiment}).

Each of the following methods (stop word removal, $L^2$ rescaling and
tf-idf weighting) transform a base bag of words matrix $C$ into a
feature matrix $X$.

\textit{Stop words removal.} Stop words are high frequency but low
information words such as ``\emph{and},'' or ``\emph{the}.''
High-frequency words have higher variance and effective weight in many
methods, often causing them to be erroneously selected as features due
to sample noise.
To deal with these nuisance words, many text-processing methods use a
fixed, hand-built stop-word list and preemptively remove all features
on that list from consideration [e.g., \citet{zhang01,ifrim08,genkin07}].
For our framework, this method generates $X$ from $C$ by ``dropping''
the columns of $C$ which correspond to a stop-word feature (while
letting $X$ take on $C$'s values exactly in the retained, nonstop-word
feature columns).

This somewhat ad hoc method does not adapt automatically to the
individual character of a given corpus and this presents many difficulties.
Stop words may be context dependent. For example, in US international
news ``\emph{united states}'' or ``\emph{country}'' seem to be high
frequency and low information. Switching to a corpus of a different
language would require new stop-word lists.
More importantly, when considering phrases instead of single words, the
stop-word list is not naturally or easily extended.\vspace*{1pt}

\textit{$L^2$-rescaled}. As an alternative, appropriately adjusting
the document vectors can act in lieu of a stop-word list by reducing
the variance and weight of high-frequency features. We use the corpus
to estimate baseline appearance rates for each feature and then adjust
the matrix $C$ by a function of these rates; see \citet{mosteller84} and
\citet{monroe08}.

We say $X$ is a {\em$L^2$-rescaled} version of $C$ if each column of $C$
is rescaled to have unit length under the $L^2$ norm, that is,
\[
L^2 \mbox{ rescaling:}\qquad x_{ij} = \frac{c_{ij}}{\sqrt{z_j}}\qquad \mbox{where } z_j \equiv\sum_{i=1}^{n}
c_{ij}^2.
\]
Under this rescaling, the more frequent a phrase, the lower its weight.

\textit{tf-idf weighting.} An alternative rescaling comes from the
popular tf-idf heuristic [\citet{salton88,salton1991dat}], which
attempts to de-emphasize commonly occurring terms while also accounting
for each document's length.
$X$ is a \emph{tf-idf weighted} version of $C$ if
\[
\mbox{tf-idf:}\qquad x_{ij}:= \frac{c_{ij}}{q_{i}} \log \biggl(
\frac
{n}{d_j} \biggr),
\]
where $q_{i} \equiv\sum_{j=1}^p c_{ij}$ is the sum of the counts of
all key phrases in document $i$ and $d_j \equiv\sum_{i=1}^n \mathds
{1}\{ c_{ij} > 0 \}$ is the number of documents in which term $j$
appears at least once.


\subsection{Feature selection methods}
\label{sec:selection}

Many prediction approaches yield models that give each feature a
nonzero weight.
We, however, want to ensure that the number of phrases selected is
small so the researcher can easily read and evaluate the entire summary
and compare it to others.
These summaries can even be automatically translated to other languages
to more easily compare foreign language news sources [\citet{xinyuetal11}].

Given the feature matrix $X$ and document labels $y$ for a topic, we
extract phrases corresponding to columns of $X$ to constitute the final summary.
We seek a subset of phrases $\mathcal{K} \subseteq\mathcal{J}$ with
cardinality as close as possible to, but no larger than, a~target~$k$,
the desired summary length. We typically use $k=15$ phrases, but 30 or
50 might also be desirable depending on the context.
We require selected phrases to be \emph{distinct}, meaning that we
do not count sub-phrases. For example, ``united states'' and ``united''
are both selected, we drop ``united.''

The constraint of short summaries renders the summarization problem a
sparse feature selection problem, as studied in, for example, \citet
{foreman2003bns,lee2006nmt,yang1997csf}.
In other domains, $L^1$-regularized methods are useful for sparse model
selection; they can identify relevant features associated with some
outcome within a large set of mostly irrelevant features.
In our domain, however, there is no reasonable expectation of an
underlying ``true'' model that is sparse; we expect different phrases
to be at least somewhat relevant.
Our pursuit of a sparse model is motivated instead by a need for
results which can be described \emph{concisely}---a constraint that
crowds out consideration of complicated dense or nonlinear
classification models.
We nonetheless employ the sparse methods, hoping that they will select
only the most important features.

We examine four methods for extraction or selection, detailed below.
Two of them, Co-occurrence and Correlation Screening, are scoring
schemes where each feature is scored independently and top-scoring
features are taken as a summary.
This is similar to traditional key-phrase extraction techniques and to
other methods currently used to generate word clouds and other text
visualizations.
The other two are $L^1$-regularized least squares linear regression
(the Lasso) and logistic regression (L1LR).
Table~\ref{tab:sampleChina} displays four summaries for China in 2009,
one from each feature selector: choice matters greatly.
We systematically evaluate this differing quality with a human
validation experiment in Section~\ref{sec:humanexperiment}.

%
\begin{table}
\caption{Comparison of the four feature selection methods. Four sample
summaries of news coverage of China in 2009. (Documents labeled via
count-2 on articles, $X$ from $L^2$-rescaling.) Note increased
prevalence of stop words in first column and redundancies in second column}
\label{tab:sampleChina}
\begin{tabular*}{\textwidth}{@{\extracolsep{\fill}}lcccc@{}}
\hline
& \textbf{Co-occurrence} & \textbf{Correlation} & \textbf{L1LR} & \textbf{Lasso} \\
\hline
\phantom{0}1 & and & beijing and & asian & asian \\
\phantom{0}2 & by & beijings & beijing & beijing \\
\phantom{0}3 &contributed  & contributed & contributed &contributed \\
& research &research &research & research \\
\phantom{0}4 & for & from beijing & euna lee & exports \\
\phantom{0}5 & global & global & global & global \\
\phantom{0}6 & has & in beijing & hong kong & hong kong \\
\phantom{0}7 & hu jintao & li & jintao & jintao \\
\phantom{0}8 & in beijing & minister wen jiabao & north korea & north korea \\
\phantom{0}9 & its & president hu jintao & shanghai & shanghai \\
10 & of & prime minister wen & staterun & tibet \\
11 & that & shanghai & uighurs & uighurs \\
12 & the & the beijing & wen jiabao & wen jiabao \\
13 & to & tibet & xinhua & xinhua \\
14 & xinhua & xinhua the & & \\
15 & year & zhang & & \\
\hline
\end{tabular*}
\end{table}

\subsubsection{Co-occurrence and correlation screening}

Co-occurrence is a simple method included in our experiments as a
useful baseline. The idea is to take phrases that appear most often (or
have greatest weight) in the positively marked text as the summary.
This method is often used in tools such as newspaper charts showing the
trends of major words over a year (such as Google News Trends\footnote
{\url{http://www.google.com/trends}.}) or word or tag clouds (created
at sites such as Wordle\footnote{\url{http://www.wordle.net/}.}).
Correlation Screening selects features with the largest absolute
Pearson correlation with the topic labeling $y$.

Both methods give each phrase a relevance score $s_j$, rank the phrases
by these~$s_j$, and then take the top $k$ phrases, dropping any
sub-phrases, as the summary.
For Co-occurrence, the relevance score $s_j$ of feature $j$ for all $j
\in\mathcal{J}$ is
\[
\mbox{Co-occurrence:}\qquad s_j = \frac{1}{\# \mathcal{I}^+ } \sum
_{i \in
I^+} x_{ij},
\]
where $\mathcal{I}^+ = \{i \in\mathcal{I} | y_i = +1\}$, that is,
$s_j$ is the average weight of phrase $j$ in the positively marked examples.
If $X=C$, that is, it is not weighted, then $s_j$ is the average number
of times feature $j$ appears in $\mathcal{I}^+$ and this method selects
those phrases that appear most frequently in the positive examples.
The weighting step, however, reduces the Co-occurrence score for common
words that appear frequently in both the positive and negative examples.

For Correlation Screening, score each feature as
\[
\mbox{Correl. Screen:}\qquad s_j = \bigl| \operatorname{cor}( x_j, y )\bigr|
\\
= \biggl\llvert \frac{\sum_{i=1}^n (x_{ij}-\bar{x}_j)(y_i -\bar{y})}{ \sqrt{\sum_{i=1}^n (x_{ij}
- \bar{x}_j)^2 }\sqrt{\sum_{i=1}^n (y_{i} - \bar{y})^2}} \biggr\rrvert,
\]
where $\bar{x}_j$ and $\bar{y}$ are the mean values of feature $j$ and
the labels, respectively, across the considered documents.

\subsubsection{$L^1$-penalized methods: Lasso and L1LR}

The Lasso [\citet{tibshirani1996regression}] is an $L^1$-penalized
version of linear regression and is the first of two feature selection
methods examined in this paper that address our
model-sparsity-for-interpretability constraint explicitly, rather than
via thresholding.
Imposing an $L^1$ penalty on a least-squares problem regularizes the
vector of coefficients, allowing for optimal model fit in
high-dimensional ($p > n$) regression settings. Furthermore, $L^1$
penalties typically result in sparse feature-vectors, which is
desirable in our context.
The Lasso also takes advantage of the correlation structure of the
features to, to a certain extent, avoid selecting highly correlated terms.

The Lasso can be defined as an optimization problem:
%
\begin{equation}
\bigl(\hat\beta(\lambda),\hat\gamma\bigr):= \arg \min_{\beta,\gamma}
\sum_{i=1}^{m} \bigl\llVert y -
x_i^T\beta- \gamma\bigr\rrVert ^2 + \lambda
\sum_j |\beta_j|. \label{sparseLasso}
\end{equation}
We solve this convex optimization problem with a modified version of
the BBR algorithm [\citet{genkin07}].
The phrases corresponding to the nonzero elements of $\beta$ comprise
our summary.
The penalty term $\lambda$ governs the number of nonzero elements of
$\beta$ and would traditionally be chosen via cross-validation to
optimize some reasonable metric for prediction.
We, however, select $\lambda$ to achieve a desired prespecified summary
length, that is, a desired number of nonzero $\beta$'s.
We find $\lambda$ by a line search.

Not tuning for prediction raises concerns of serious over- or under-fitting.
Generally, in order to have short summaries, we indeed under-fit.
Additionally, since our labeling is not very accurate in general,
prediction performance might even be misleading.
The main question is whether a \emph{human-readable} signal survives
imperfect labeling and over-regularized summaries, both of which allow
for easier exploration of text.
These concerns motivate the human validation study we discuss in
Section~\ref{sec:humanexperiment}.

Similar to the Lasso, $L^1$-penalized logistic regression (L1LR) is
typically used to obtain a sparse feature set for predicting the
log-odds of an outcome variable being either $+1$ or $-1$. It is widely
studied in the classification literature, including text classification
[see \citet{genkin07,ifrim08,zhang01}].
For an overview of the Lasso, $L^1$-penalized logistic regression and
other sparse methods see, for example, \citet{HTF03}. For details of
our implementation along with further discussion, see \citet{techreport}.

Co-occurrence, correlation screening and the Lasso are all related.
The Co-occurrence score $s_j$ can be seen as the average count (or
weighted count for a reweighted feature matrix) of phrase $j$ in the
positively marked examples, denoted as $\hat E(x_{j}|y= +1)$.
Correlation Screening is related but slightly different; calculations
show that $\operatorname{cov}(x_j,y)$ is proportional to $\hat E(x_{j}|y= +1) - \hat
E(x_{j}|y= -1)$, and hence is the difference between the positive and
negative examples [see \citet{techreport} for details].
Both Co-occurrence and Correlation Screening methods are greedy procedures.
Since the Lasso can be solved via $e$-L2boosting [\citet{zhao2007stagewise}],
the Lasso procedure can also be interpreted as greedy.
It is an iterative correlation search procedure---the first step is to
get the word/phrase with the highest correlation; then we modify the
labels to remove the influence of this word/phrase and then get the
highest correlated word/phrase with this modified label vector and so
on and so forth.

The primary advantages of Co-occurrence and Correlation Screening are
that they are fast, scalable and easily distributed across multiple
cores for parallel processing.
Unfortunately, as they score each feature independently from the
others, they cannot take advantage of any dependence between features
to aid summarization.
The Lasso and L1LR can, to a certain extent.
The down side is that the sparse methods are more computationally
intensive than Co-occurence and Correlation Screening.
However, this could be mitigated by, for example, moving to a parallel
computing environment or doing clever preprocessing such as safe
feature elimination [\citet{elghaoui11}].
For our current implementation (which is our modified form of the BBR
algorithm [\citet{genkin07}]), we timed the Lasso as being currently
about 9 times and L1LR more than 100 times slower than the baseline
Co-occurrence. See Table~\ref{tab:runningtime}.

%
\begin{table}
\caption{Computational speed chart. Average running times for the four
feature selection methods over all subjects considered. Second column
includes time to generate $y$ and adjust $X$. Final column is
percentage increase in total time over Co-occurrence, the baseline method}
\label{tab:runningtime}
\begin{tabular*}{\textwidth}{@{\extracolsep{\fill}}lccc@{}}
\hline
&\textbf{Phrase selection (sec)} &\textbf{Total time (sec)} &\textbf{Percent increase} \\
\hline
Co-occurrence & \phantom{00}1.0 & \phantom{0}20.3 & \\
Correlation screen & \phantom{00}1.0 & \phantom{0}20.3 & \phantom{000}0\% \\
The Lasso & \phantom{00}9.3 & \phantom{0}28.7 & \phantom{0}$+$41\% \\
L1LR & 104.9 & 124.2 & $+$511\% \\
\hline
\end{tabular*}
\end{table}

\section{The human validation survey}
\label{sec:humanexperiment}

Consider the four sample summaries on Table \ref{tab:sampleChina}. These particular
summaries came from a specific combination of
choices for the reweighting ($L^2$-rescaling), labeling (count-2)
and feature selection steps (co-occurrence, correlation, L1LR and the Lasso).
But are these summaries better or worse than the summaries from a
different summarizer
with another specific combination?

Comparing the efficacy of different summarizers requires systematic evaluation.
To do this, many researchers use corpora with existing summaries, such
as human-encoded key phrases in academic journals such as in \citet
{franketal1999} or baseline human-generated summaries such as the
TIPSTER data set used in \citet{netoetal2002}.
We, however, give a single summary for many documents, and so we cannot
use an annotated evaluation corpus or summaries of individual documents.

Alternatively, numerical measures such as prediction accuracy or model
fit might be used to compare different methods.
However, the major purpose of text summarization is to help humans
gather information, so the quality of summarization should be
compared to human understanding based on the same text.
While we hypothesize that prediction accuracy or model fit should
correlate with summary quality as measured by human evaluation to a
certain extent, there are no results to demonstrate this.
Indeed, some research indicates that the correlation between good model
fit and good summary quality may be absent, or even negative, in some
experiments [\citet{mir,chang09}].


In this section, therefore, we design and conduct a study where humans
assess summary quality.
We compare our four feature selection methods under different
text-segmenting, labeling and weighting choices in a crossed and
randomized experiment.
Nonexperts read both original documents and our summaries in the experiment
and judge the quality and relevance of the output. Even though we
expect individuals' judgements to vary, we can average the responses
across a collection of respondents and thus
get a measure of overall, generally shared opinion.

\subsection{Human survey through a multiple-choice questionnaire}

We carried out our survey in conjunction with the XLab, a UC Berkeley
lab dedicated to helping
researchers conduct human experiments. We recruited 36 respondents
(undergraduates at a major university) from the lab's respondent pool
via a generic, nonspecific message stating that there was a study that
would take up to one hour of time.
For our investigation we used the International Section of the New York
Times for 2009. See our first case study in Section~\ref{sec:casestudy}
for details on this data set.

We evaluated 96 different summarizers built from different combinations
along the following four dimensions:
\begin{longlist}[\quad]
\item[\textit{Document unit}:] When building $C$, the document units
corresponding to the matrix rows may be either (1) full articles or
(2) the individual paragraphs in those articles.
\item[\textit{Labeling}:] Documents can be labeled according to the rules
(described in the preceding section) (1) count-1, (2) count-2,
(3) count-3, (4) hcount-2 or (5) \mbox{hcount-3}.
\item[\textit{Rescaling}:] Matrix $X$ can be built from $C$ via (1) stop-word
removal, (2) $L^2$ rescaling or (3) tf-idf weighting.
\item[\textit{Feature selection}:] Data $(X,y)$ can be reduced to a summary using
(1) Co-occurrence, (2) Correlation Screening, (3) the Lasso or (4) L1LR.
\end{longlist}
Together, for any given query, there exist $2 \times 5 \times 3 \times
4 = 120$ CCS summary methods available. We dropped count-3 and
Hcount-3 for paragraphs giving 96 tested.

We applied each summarizer to the set of all articles in the New York
Times International Section from 2009 for 15 different countries of
interest. These countries are
\textit{China, Iran, Iraq, Afghanistan, Israel, Pakistan, Russia,
France, India, Germany, Japan, Mexico, South Korea, Egypt and Turkey}.
The frequency of appearance in our data for these countries can be
found in Table~6 of \citet{techreport}.
We then compared the efficacy of these combinations by having respondents
assess (through answering multiple-choice questions) the quality of
the summaries generated by each summarizer.

For our survey, paid respondents were convened in a large room of
kiosks where they assessed a series of summaries and articles presented
in 6 blocks of 8 questions each.
Each block considered a single (randomly selected) topic from our list
of 15. Within a block, respondents were first asked to read four
articles and rate their relevance to the specified topic. Respondents
were then asked to read and rate four summaries of that topic randomly
chosen from the subject's library of 96. Respondents could not go back
to previous questions.

Only the first 120 words of each article were shown. Consultation with
journalists suggests this would not have a detrimental impact on
content presented, as a traditional newspaper article's ``inverted
pyramid'' structure moves from the most important information to more
minute details as it progresses [\citet{Pottker03}].
All respondents finished their full survey, and fewer than 1\% of the
questions were skipped.
Time to completion ranged from 14 to 41 minutes, with a mean completion
time of 27 minutes.
See \citet{techreport} for further details and for the wording of the survey.

\subsection{Human survey results}
\label{sec:humanvalidresults}

\begin{figure}

\includegraphics{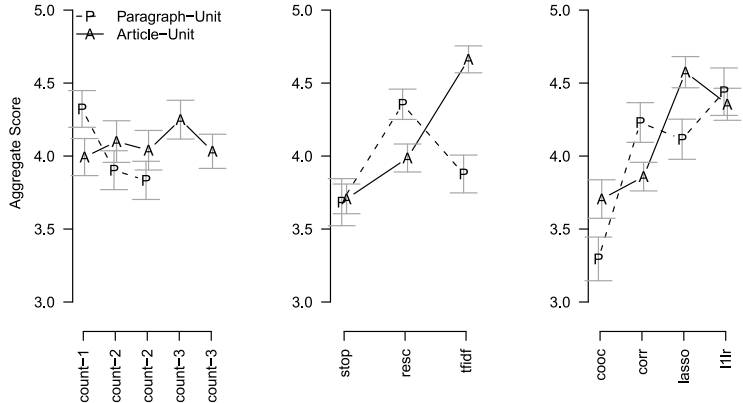}

\caption{Aggregate results. Outcome is aggregate score based on the raw data.
There are major differences between article-unit analysis and
paragraph-unit analysis when considering the impact of choices
in preprocessing. Error bars are $\pm1$ unadjusted SE based only on
subset of scores at given factor combinations.}
\label{fig:interactArtPar}
\end{figure}

We primarily examined an aggregate ``quality'' score, taken as the mean
of the assessed Content, Relevance and Redundancy of the summaries.
Figure~\ref{fig:interactArtPar} shows the raw mean aggregate outcomes
for the article-unit and paragraph-unit data.
The rightmost plot suggests that the Lasso and L1LR performed better
overall than Co-Occurrence and Correlation Screen.

We analyze the data by fitting the respondents' responses to the
summarizer characteristics using linear regression, although all plots
here show raw, unadjusted data. The adjusted plots show similar trends.
The full model includes terms for respondent, subject, unit type,
rescaling used, labeling used and feature selector used, as well as all
interaction terms for the latter four factors.

In all models, there are large respondent and topic effects. Some
topics were more easily summarized than others, and some respondents
more critical than others. Interactions between the four summarization
method factors are (unsurprisingly) present ($\mathit{df}=33$, $F=4.14$, $\log_10 p
\approx-13$ under ANOVA).
There are significant three-way interactions between unit,
feature-selector and rescaling ($p \approx0.03$) and labeling,
feature-selector and rescaling ($p \approx0.03$).
Interaction plots (Figure~\ref{fig:interactArtPar}) suggest that the
sizes of these interactions are large, making interpretation of the
marginal differences for each factor potentially misleading. Table~\ref{tab:testing} shows all significant two-way interactions and main
effects for the full model, as well as for models run on the
article-unit and paragraph-unit data separately.

As the unit of analysis heavily interacts with the other three factors,
we conduct further analysis of the article-unit and paragraph-unit data
separately.
The article-unit analysis is below.
The paragraph-unit analysis, not shown, is summarized in
Section~\ref{sub:discussion}'s discussion on overall findings.

\begin{figure}

\includegraphics{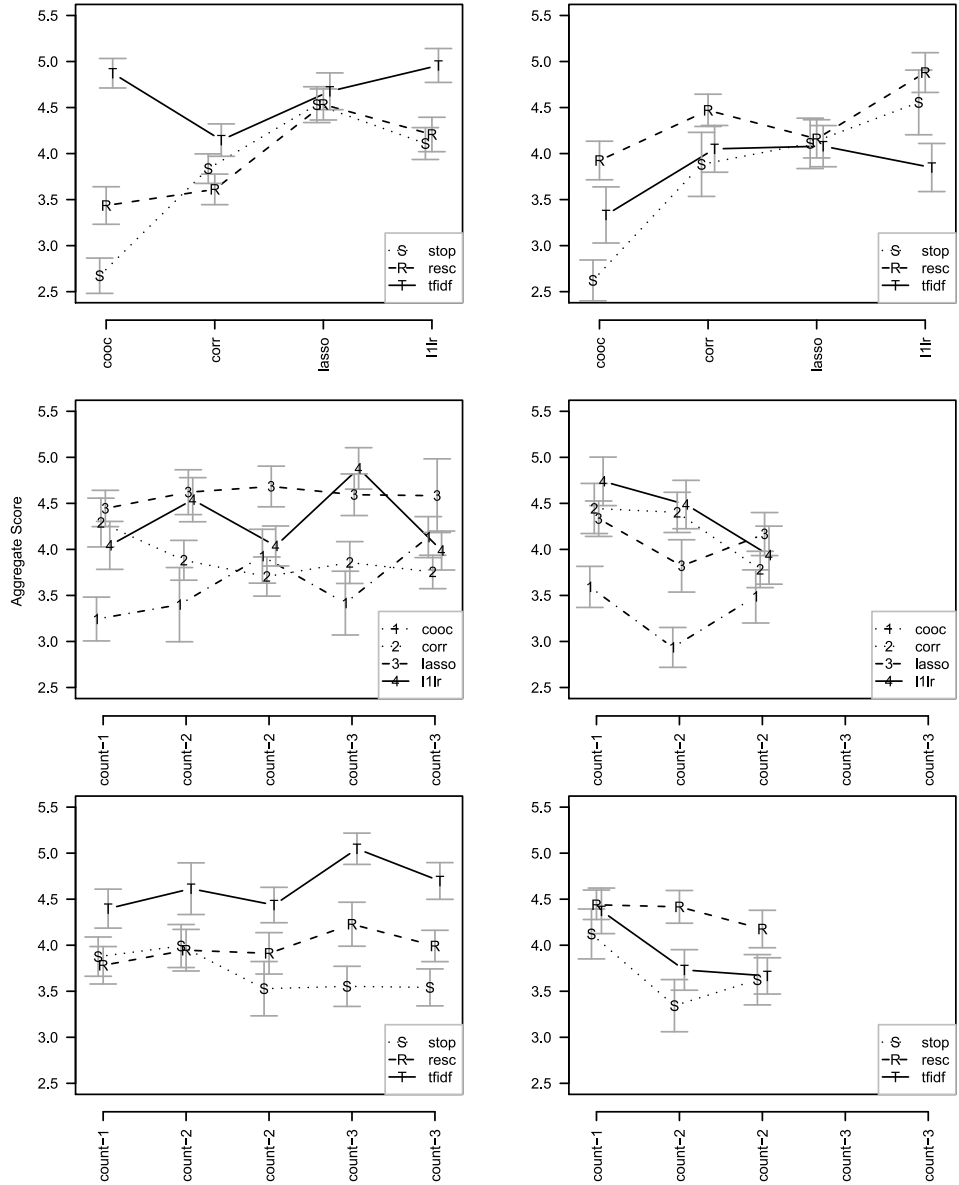}

\caption{Aggregate quality plots.
Pairwise interactions of feature selector, labeling and rescaling
technique. Left-hand side is for article-unit
summarizers, right for paragraph-unit. See testing results for which
interactions are significant.}
\label{fig:interactAgg}
\end{figure}

\begin{table}[b]
\caption{Main effects and interactions of factors. Main effects along
diagonal in bold. A number denotes a significant main effect or
pairwise interaction for aggregate scores and is the (rounded) base-10
log of the $p$-value. ``.'' denotes lack of significance at the 0.05
level. ``All data'' is all data in a single model without third- and
fourth-order interactions. ``Article-unit'' and ``paragraph-unit''
indicate models run on only those data for summarizers operating at
that level of granularity}
\label{tab:testing}
\begin{tabular*}{\textwidth}{@{\extracolsep{\fill}}lcccccccccc@{}}
\hline
& \multicolumn{4}{c}{\textbf{All data}} & \multicolumn{3}{c}{\textbf{Article-unit}} &
\multicolumn{3}{c@{}}{\textbf{Paragraph-unit}}\\[-6pt]
& \multicolumn{4}{c}{\hrulefill} & \multicolumn{3}{c}{\hrulefill} &
\multicolumn{3}{c@{}}{\hrulefill}\\
\textbf{Factor} & \textbf{Unit} & \textbf{Feat.} & \textbf{Lab.} & \textbf{Resc.} &
\textbf{Feat.} & \textbf{Lab.} & \textbf{Resc.} & \textbf{Feat.} & \textbf{Lab.} &
\textbf{Resc.} \\
\hline
Unit & \textbf{.} & \phantom{0}$-2$ &. & \phantom{0}$-7$ & & & & & \\
Feat. select & & \textbf{$\bolds{-}$17} &. & $-10$ & \textbf{$\bolds{-}$10} &.
& \phantom{0}$-8$ &
\textbf{$\bolds{-}$7} &. & $-2$\\
Labeling & & & \textbf{.} &. & & \textbf{.} &. & & \textbf{$\bolds{-}$2} &. \\
Rescaling & & & & \textbf{$\bolds{-}$14} & & & \textbf{$\bolds{-}$15} & & &
\textbf{$\bolds{-}$3}\\
\hline
\end{tabular*}
\end{table}

\begin{table}[b]
\caption{Quality of feature selectors. This table compares the
significance of the separation of the feature selection methods on the
margin. Order is always from lowest to highest estimated quality. A
``$<$'' denotes a significant separation. All $p$-values corrected for
multiple pairwise testing. The last seven lines are lower power due to
subsetting the data}
\label{tab:quality}
\begin{tabular*}{\textwidth}{@{\extracolsep{\fill}}lcc@{}}
\hline
\textbf{Data included} & \textbf{Order (article)} & \textbf{Order (paragraph)} \\
\hline
All & cooc, corr $<$ L1LR, Lasso & cooc $<$ corr, Lasso, L1LR \\
& stop $<$ resc $<$ tf-idf & tfidf, stop $<$ resc \\[3pt]
tf-idf only & no differences & no differences\\
$L^2$ only & cooc $<$ L1LR, Lasso; corr $<$ Lasso & no differences \\
stop only & cooc $<$ corr, L1LR, Lasso; corr $<$ Lasso & cooc $<$
Lasso, L1LR \\[3pt]
cooc only & stop $<$ resc $<$ tf-idf & stop $<$ resc \\
corr only & stop $<$ tf-idf & no differences \\
Lasso only & no differences & no differences \\
L1LR only & no differences & tf-idf $<$ resc \\
\hline
\end{tabular*}
\end{table}

\textit{Article-unit analysis.} 
The left column of Figure~\ref{fig:interactAgg} shows, for the
article-unit data, plots of the three two-way interactions between
feature selector, labeling scheme and rescaling method. There is a
strong interaction between the rescaling and feature-selection method
($\mathit{df}=6$, $F=8.07$, $\log p \approx-8$, top-left plot), and no evidence of
a labeling by feature-selection interaction or a labeling by rescaling
interaction.
Model-adjusted plots (not shown) akin to Figure~\ref{fig:interactAgg}
do not differ substantially in character. Table~\ref{tab:testing} shows all
significant ($\alpha= 0.05$) main effects and pairwise interactions.

The Lasso is the most consistent method, maintaining high scores under
almost all combinations of the other two factors. In Figure~\ref{fig:interactAgg}, note how the Lasso has a tight cluster of means
regardless of the rescaling method used in the top-left plot and how
the Lasso's outcomes are high and consistent across all labeling in the
middle-left plot.
Though L1LR or Co-occurrence may be slightly superior to the Lasso when
coupled with tf-idf, they are not greatly so, and, regardless, both
these methods seem fragile, varying a great deal in their outcomes
based on the text preprocessing choices.

Validating its long history of use, tf-idf seems to be the best overall
rescaling technique, consistently coming out ahead regardless of choice
of labeling or feature-selection method.
Note how its curve is higher than the rescaling and stop-word curves in
both the top- and bottom-left plots in Figure~\ref{fig:interactAgg}.
Weighting by tf-idf brings otherwise poor feature selectors up to the
level of the better selectors.

We partially ordered the levels of each factor by overall (marginal)
impact on summary quality.
For each factor, we fit a model with no interaction terms for the
factor of interest to get its marginal performance and, within this
model, performed pairwise testing for all levels of the factor,
adjusting the resulting $p$-values to control familywise error rate
with Tukey's honest significant difference to address the
multiple-testing problem within each factor.
These calculations showed which choices are overall good performers
(ignoring interactions).
See Table~\ref{tab:quality} for the resulting rankings.
Co-occurrence and Correlation Screening performed significantly worse
than L1LR and the Lasso (correlation vs. L1LR gives $t = 3.46$, $p < 0.05$).
The labeling method options are indistinguishable.
The rescaling method options are ordered with tf-idf significantly
better than rescaling ($t=5.08$, $\log p \approx-4$), which in turn is
better than stop-word removal ($t=2.45$, $p < 0.05$).

\textit{Discussion.}
\label{sub:discussion}
Comparing the performance of the feature selectors is difficult due to
the different nature of interactions for paragraph and article units.
That said, the Lasso consistently performed well. When building $C$ at
the article-unit level, Lasso was a top performer. For the
paragraph-unit it did better than most but was not as definitively
superior. L1LR, if appropriately staged, also performs well.

Simple methods such as Co-occurrence are sensitive to the choice of
weighting method and, generally speaking, it is hard to know what
weighting is best for a given corpus.
This sensitivity is shared by L1LR. Under the Lasso, however, these
decisions seem unimportant regardless of unit size.
We therefore recommend using the Lasso, as it is far less sensitive to
the choice of weights.

\textit{A note on tf-idf and $L^2$ rescaling.}
The main difference between the paragraph-unit and article-unit data is
that tf-idf is a poor choice of rescaling and $L^2$-rescaling is the
best choice for paragraph-unit.
We conducted a further investigation to understand why this was the
case and found that any given stop word will appear in most articles,
due to the articles' lengths, which under tf-idf will result in very
small weights.
Low weight words are hard to select and, thus, those terms are dropped.
For the paragraph-unit level, however, the weights are not shrunk by
nearly as much since many paragraphs will not have any particular
low-content word. (For example, prepositions like ``among'' or ``with.'')

The $L^2$ recalling, however, maintains the low weights, as the weight
basically depends on total counts across the corpus.
If one makes histograms of these weights (not shown), this shift is
readily apparent.
For short units of text, $L^2$ rescaling is a stronger choice since it
is not sensitive to document length.
Of course, the Lasso makes these decisions less relevant.

\section{Case studies}
\label{sec:casestudy}
Here we illustrate our CCS tool by conducting two example analyses that
demonstrate how researchers can explore corpora, collect evidence for
existing theories and generate new theories.
That is, we here attempt to meaningfully connect our methodology to
actual practice, an orientation to research argued for in, for example,
\citet{Wagstaff:2012tr}.

Given the validation of the human reader survey, we restrict CCS to use
the Lasso with $L^2$ regularization over full articles with a
``count-1'' rule, a combination determined most effective overall by
the human experiment.
In the first study, we conduct an analysis of how Egypt was covered by
the international section of the New York Times throughout the Arab Spring.
In the second, we compare the headlines of the New York Times to those
of the Wall Street Journal on the topics of ``energy.''

\subsection{Egypt as covered by the international section of the New
York Times}

We here investigate how Egypt was framed across time in the
International Section of the New York Times from the beginning of 2009
through July, 2012.\footnote{Clavier and Barnesmoore are conducting a
larger study on the topic.}
Through this analysis, we hope to illuminate both consistent and
changing trends in the coverage of Egypt as well as the impact of
different stages of the Arab Spring on how Egypt was editorially framed.
Though of course there are a myriad of frames and narratives, we
selected a few of the most influential, recognizable and contextually
established narratives to remain within the scope of this paper and to
provide a basic overview of possible applications for these tools in
the analysis of media representation.

This study demonstrates how CCS can be used to examine how the framing
of countries and political entities can evolve throughout the
progression of political situations such as revolutions and elections.
We show that our tool can also help determine the more macro frames of
narration that structure coverage of a region.
We argue the findings from our tool allow an analyst to better
understand the basic logic of reporting for a region and how events
such as uprisings and key elections impact that coverage.

Articles were scraped from the New York Times' RSS feed,\footnote
{\url{feed://feeds.nytimes.com/nyt/rss/World}.} and the HTML markup
was stripped from the text.
We obtained 35,444 articles.
The New York Times, upon occasion, will edit an article and repost it
under a different headline and link; these multiple versions of the
articles remain in the data set.
By looking for similar articles, as measured by a small angle between
their feature vectors in the document-term matrix $C$, we estimate that
around 4--5\% have near-duplicates.

The number of paragraphs in an article ranges from 1 to 38.
Typical articles\footnote{See, for example, \texttt{%
\href{http://www.nytimes.com/2011/03/04/world/americas/04mexico.html}{http://www.nytimes.com/2011/03/04/world/americas/04mexico.}
\href{http://www.nytimes.com/2011/03/04/world/americas/04mexico.html}{html}}.}
have
about 16 paragraphs [with an Inter-Quartile Range (IQR) of 11 to 20 paragraphs].
However, about 15\% of the articles, the ``World Briefing'' articles,
are a special variety that contain only one long paragraph.\footnote
{See, for example, \texttt{%
\href{http://www.nytimes.com/2011/03/03/world/americas/03briefs-cuba.html}{http://www.nytimes.com/2011/03/03/world/americas/03briefs-}
\href{http://www.nytimes.com/2011/03/03/world/americas/03briefs-cuba.html}{cuba.html}}.}
Among the more typical, non-``World Briefing'' articles, the
distribution of article length as number of paragraphs is bell-shaped
and unimodal.
Longer articles, with a median length of 664 words, have much shorter
paragraphs (median of 38 words), generally, than the ``Word Briefing''
single-paragraph articles (median of 87 words).

In the early 1990s, Entman posited that our learning of the world is
built on frames which he defines as ``information-processing schemata''
that ``operate by selecting and highlighting some features of reality
while omitting others'' [\citet{entman1993}, page 53].
Media studies incorporate these definitions by showing the role of the
media in creating these frames, stating, for example, that ``through
choice and language and repetition of certain story schemas,'' the
media ``organizes and frames reality in distinctive ways'' [\citet
{mcleodetal1991}].
Following \citet{goffman1974}, we agree that the analysts' task
therefore is to identify frames in media discourse within the
understanding that media framing, under the guise of informing, can
deliberately influence public opinion.
Indeed, most of the literature on framing and subsequent agenda-setting
literature argues that frames are purposely created.
According to Entman, ``to frame is to select some aspects of a
perceived reality and make them more salient in a communicating text,
in such a way as to promote a particular problem, moral evaluation,
and/or treatment recommendation'' [\citet{entman2004}].

In terms of portrayal of other countries, frames tend to be easy to
observe, as popular news media tend to establish simplified dichotomies
of ``we'' versus ``other'' and they classify data under those two
categories, often outlined as mirror images of positive attributes
versus negative ones [\citet{kiousisetal2008,kunczik2000}].
Given that frames in the media center around repeated, and often
simplified, elements, our tools seem to naturally lend themselves to
the extraction of a frame's ``fingerprint.''
At core, our methods extract relevant phrases that are often repeated
in conjunction with a topic of interest.
These phrases, when read as news, arguably build links in readers'
minds to the topic and thus contribute to the formation and
solidification of how the topic is framed.

%
\begin{table}[b]
\caption{Overview of the NYT windows for the Egypt summary. Columns
encode stats during each period: time period name, start and stop
dates, total number of articles, number of articles about Egypt, number
of Egypt articles per week, and Egypt article volume as a percentage of
total volume}
\label{tab:egyptsummary}
\begin{tabular*}{\textwidth}{@{\extracolsep{\fill}}lcccccc@{}}
\hline
\textbf{Period} & \textbf{Start} & \textbf{Stop}& \textbf{\#Art.} &
\textbf{\#Egypt} & \textbf{Eg./Wk} & \textbf{\%Egypt} \\
\hline
2009 & 01-Jan-09 & 31-Dec-09 & 9560 & 485 & \phantom{0}9.3 & \phantom{0}5.1 \\
2010 & 01-Jan-10 & 31-Oct-10 & 8519 & 312 & \phantom{0}7.2 & \phantom{0}3.7 \\
Before uprisings & 01-Nov-10 & 16-Dec-10 & 1272 & \phantom{0}62 & \phantom{0}9.6 &
\phantom{0}4.9 \\
Revolution & 17-Dec-10 & 01-Mar-11 & 2098 & 428 & 40.5 & 20.4 \\
Post Mumbarak & 01-Mar-11 & 31-Oct-11 & 6896 & 767 & 22.0 & 11.1 \\
Parl elections & 01-Nov-11 & 30-Jan-12 & 2476 & 219 & 17.0 & \phantom{0}8.8 \\
Post elections & 01-Feb-12 & 01-Jul-12 & 3585 & 249 & 11.5 & \phantom{0}6.9 \\[3pt]
Whole corpus & 01-Jan-09 & 01-Jul-12 & 34,406 & 2522 & 13.8 & 7.3 \\
\hline
\end{tabular*}
\end{table}

To capture the evolving frames of Egypt and elections across time, we
generated several sequences of summaries.
We summarized within specific windows of time with boundaries
determined by major political events such as the beginning of the
uprisings in Tunisia (December 16th, 2010) or Egyptian parliamentary
elections (February 1st, 2012). See Table~\ref{tab:egyptsummary}.
We present summaries of different periods of time; an alternate
approach would be to attempt to link articles and present a graph of
relationships. See, for example, \citet{shahaf2012trains} or \citet{ELDPSB:11}.

We first generated CCS summaries (using the Lasso with $L^2$ rescaling
over full-article document units) comparing all articles mentioning
Egypt to all other articles.
We subsequently compared Egypt vs. the other articles within only those
articles that also contained variants of ``election'' and examined
other Arab countries (e.g., Tunisia) as well as phrases such as
``arab'' and ``arab spring.'' This process generated several graphical
displays of summaries, all examining different facets of news coverage
from the NYT.

\begin{figure}
\includegraphics{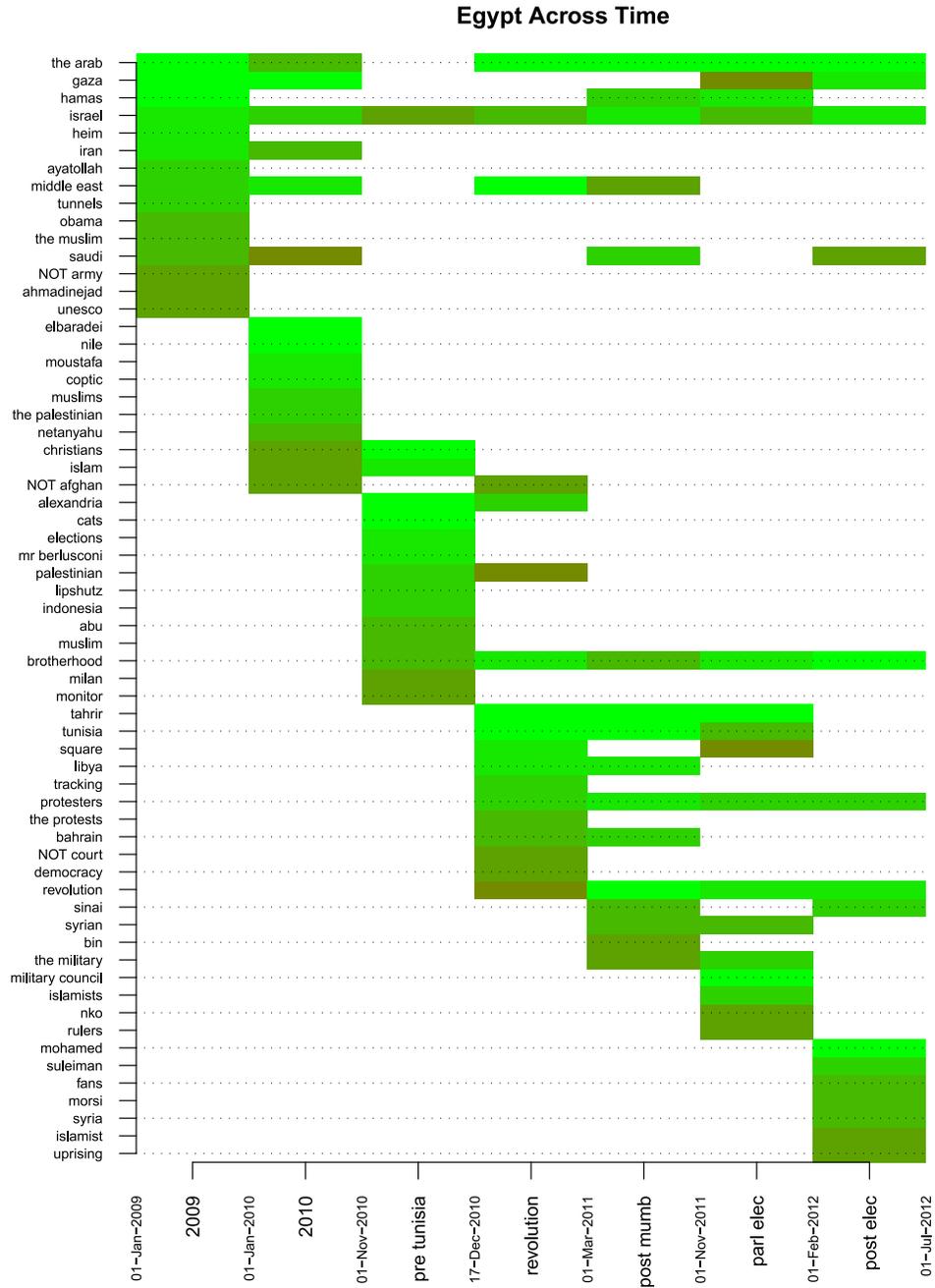}
\caption{Framing of Egypt. Columns correspond to
prespecified windows of time.}
\label{fig:egypt}
\end{figure}

For an example see Figure~\ref{fig:egypt}, which shows the overall
framing of Egypt across time.
We identified articles as Egypt-related if they contained any of \emph
{egypt, egypts, egyptian, egyptians, cairo, mubarak} (the \emph
{count-1} rule).
We analyzed at the article level and used the Lasso with tf-idf regularization.
After looking at the first list, we removed ``arab'' and ``hosni'' as
uninformative and re-ran our summarizer to focus the summary on more
content-relevant phrases.
Such an iterative process is, we argued, a more natural and principled
way of discovering and eliminating low-content features; in this case
``Hosni'' is Mubarak's first name, and ``arab'' tends to show up in
articles in this region as compared to other regions.
Neither of these words would be found on any typical stop-word list.

From Figure~\ref{fig:egypt}, and others similar to it, we can consider
consistent and changing trends in the coverage of Egypt as well as the
impact of different stages of the Arab Spring on how Egypt was framed.
We then sampled text fragments and sentences containing these phrases
from the corpus to ensure we were interpreting them correctly.
For example, ``the arab'' in 2009 typically (but not always) appears
before ``world,'' as in, for example, ``mostly from THE ARAB world.''
We now give an overview of the resulting analysis.

\textit{Pre-Arab spring \textup{(}columns \textup{1, 2} and \textup{3)}.} The summaries, shown
as the first three columns of Figure~\ref{fig:egypt}, are for 2009,
most of 2010 and for the time just prior to the uprisings in Tunisia.
Coverage of the Arab world prior to the Arab Spring is dominated by
concern for Israel and narratives concerning the ``War on Terror.''
Note the appearance of ``Israel,'' ``Hamas,'' ``Gaza'' and ``Palestinian.''
There are two probable reasons for the appearance of these words.
First, Israel bombed Egypt in 2009.
Second, following the Camp David accords of 1979, the United State's
political, economic and military strategies within the MENA region
became reliant on sustaining these accords.
And, indeed, the Mubarak regime sustained this treaty in the face of
mass opposition by the Egyptian people.
Overall, we see Egypt as being covered in the context of its
connection of Israel and the Israeli-Palestinian conflict.

We also see, for the period just prior to the uprisings, ``cats'' and ``milan.''
These phrases are overall rare words that happened to appear at
disproportionate rates in the positively marked articles and are thus
selected as indicative.
This can happen when there are few positive examples (only 62 in this
time span) in an analysis.

\textit{Arab spring \textup{(}columns \textup{4, 5} and \textup{6)}.}
We divided the Arab Spring into three rough periods: the initial
revolution during the late months of 2010 (column 4), the time just
after the fall of President Mumbarak through 2011 (column 5), and the
time leading up to the parliamentary election at the end of 2011 into
2012 (column 6), at which point a nominal government had been established.

Throughout this time we see a shift in coverage, most obviously
indicated by the appearance of the words ``protests,'' ``protesters''
and ``revolution.''
``The arab,'' which indicated either ``the arab world'' or ``the arab
league'' before, now indicates ``the arab world'' or ``the arab
spring'' (as found by examining text snippets containing the found
summary phrases).
We see that US foreign policy imperatives retain their importance as
shown by the continued appearance of ``Israel,'' ``Hamas'' and ``Gaza.''

Note the entrance of discussion concerning the military and military
councils (e.g., ``the military'' and ``military council'') in Egyptian
coverage as elections approach.
The heightened appearance comes at a time when much discussion
concerning the elections is dominated by the Islamist nature of the
major parties running for office (see, e.g., ``islamists'' and
``[muslim] brotherhood'' in column 6 for the time just prior to the
parliamentary elections).
As the military regime in Egypt could be perceived by many in western
circles as a keystone for regional peace with Israel,
this frame of narration arguably lends a sense of stability concerning
the status quo.

\textit{After the parliamentary elections \textup{(}Column \textup{7)}.}
Following the initial elections in Egypt, the frame of Israel, Gaza and
Hamas remain, but we also see ``islamist,'' ``morsi'' and
``brotherhood,'' suggesting a developing frame of an Islamic threat to
the western domestic sphere posed by groups like the Muslim Brotherhood.
The shift comes as the western media begins to cover the elections in Egypt.
As the U.S. has supported the elections as being legitimate, the
western media is now faced with the assumption that the will of the
Egyptian public might be more fully actualized in an open democracy.
Existing American and Israeli fears of Islamic extremism mixed with the
prevalence of Islamist parties in the elections combine to form a new
frame of coverage.
This frame, however, is in many cases still dominated by the
relationship of the Islamist parties to the U.S. and its close ally Israel.

\subsection{Comparing the New York Times to the Wall Street Journal}
\label{sec:nyt_wsj}

In our second case study, we, as readers of the Wall Street Journal
(WSJ) and the New York Times (NYT),
use CCS to understand the differences and similarities of these two
major newspapers
across time. We focus on headlines. As headlines are quite short, we,
based on the human experiment results, used the Lasso with
$L^2$-rescaling and no stop-word removal.
Our data set consists of 289,497 headlines from the New York Times and
284,042 headlines from the Wall Street Journal, scraped from their RSS
feeds for four years, from Jan 1, 2008 through the end of 2011.

As a first exploratory step, we labeled NYT headlines as positive
examples and WSJ headlines as negative examples and applied CCS.
The initial results gave phrases such as ``sports,'' ``review'' and
``arts'' as indicating a headline being from the NYT.
Exploration of the raw data revealed that the NYT precedes many
headlines with a department name, for example, ``arts briefly,'' giving
this result. However, other phrases, for example, ``for'' and ``of,''
also repeatedly appear in the summaries as being indicative of the NYT.
This, coupled with the fact that very few phrases indicated the WSJ,
suggests that the NYT has a more identifiable ``signal'' for
classification, that is, a more distinctive headline style.
For further content-focused investigations we then dropped these
department-related words and phrases (e.g., sports, review, etc.) as
potential features.

We then conducted a content-focused analysis to compare the NYT and WSJ
with respect to how they cover energy, as represented by headlines
containing general words such as \emph{oil, solar, gas, energy} and
\emph{electricity}. 6605 (2.3\%) of the WSJ had headlines containing
these words, while 2462 (0.9\%) of the NYT's headlines contained these
terms. See Table~\ref{tab:energy}.
We actually investigated differently broad interpretations of this topic.
One version included \emph{energy} only, and another included words such
as \emph{oil, natural gas, solar}.
Also, with an iterative process we can conduct an informal ``keyword
expansion'' to refine the representation of their topic of interest in
the context of the corpus being examined by updating the labeling process.
For example, we here included ``natural'' as a keyword after seeing it
prominently in connection with ``energy'' as a first pass.

\begin{figure}

\includegraphics{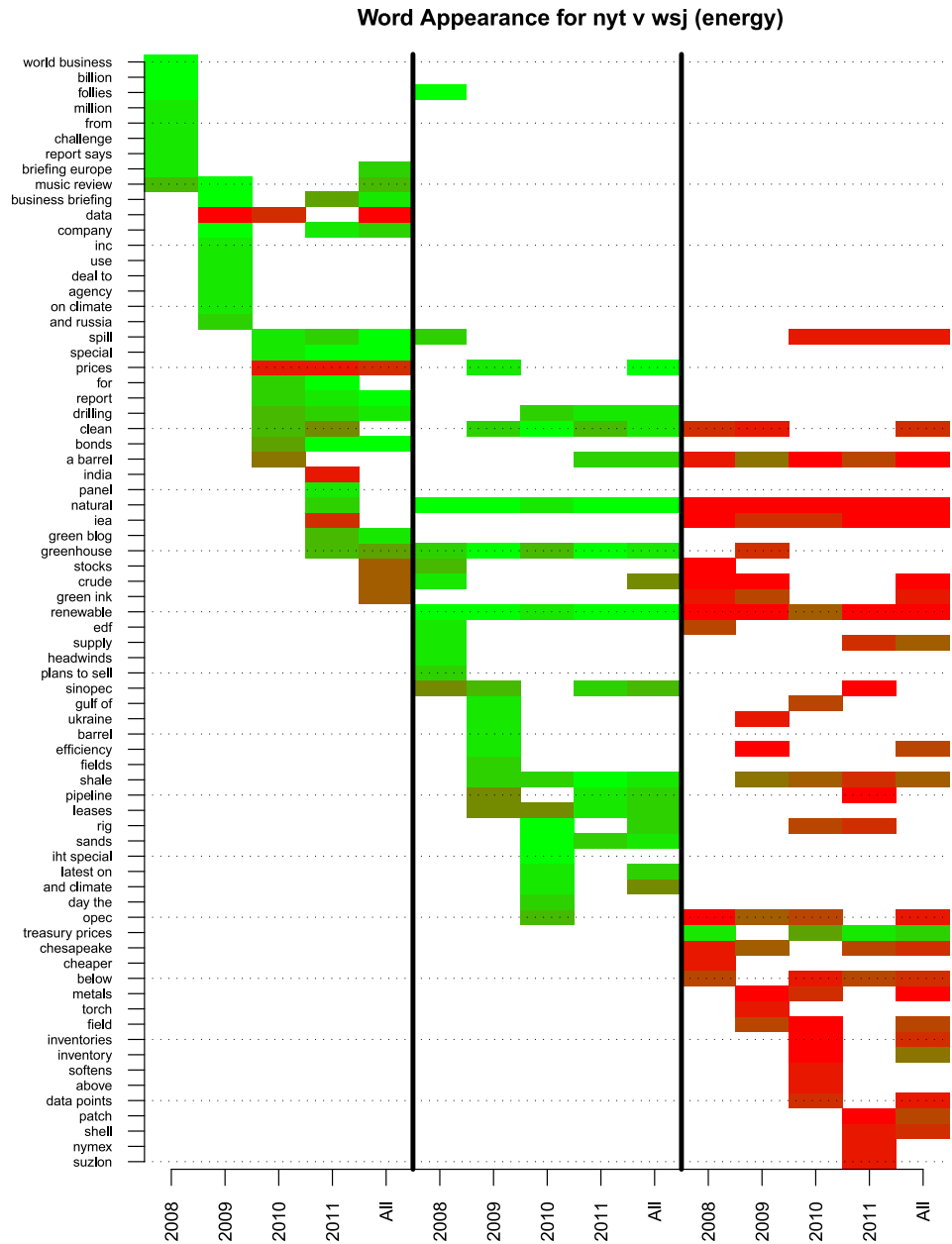}

\caption{The NYT vs. the WSJ with regards to energy.
First 5 columns are the ``between'' comparison of the
NYT vs. the WSJ. Second 5 are an internal ``within''
comparison of energy to nonenergy within the NYT.
Third set are the same for the WSJ. Red indicates
WSJ and green NYT. Within each set, columns correspond
to 2008, 2009, 2010 and 2011, respectively. ``All''
is all four years combined.}
\label{fig:nytwsjenergy}
\end{figure}

For a first summary, we did a head-to-head (or ``between-source'')
comparison as follows: we first dropped all headlines that did not
mention any of the energy-related terms. We then labeled NYT
energy-related headlines as $+1$ and WSJ energy-related headlines as
$-1$ and applied CCS.
This gave \emph{data, prices, stocks, green ink and crude} as being in
the WSJ's frame and \emph{spill, greenhouse}, \emph{world business} and
\emph{music review} as being the NYT's. See Figure~\ref{fig:nytwsjenergy}.
These latter two phrases are after several similar terms had already
been removed.
``World business'' is a department label for articles about
international affairs, and its appearance connects coverage of energy
with international news.
``Music review'' is due to 17 music review articles using ``energy'' in
headlines such as ``energy abounds released by a flurry of beats'' or
``molding sound to behave like a solar eclipse.''
A head-to-head comparison will capture stylistic differences between
the corpora as well as differences in what content is covered.


%
\begin{table}
\caption{Summary of headlines for energy investigation}
\label{tab:energy}
\begin{tabular*}{\textwidth}{@{\extracolsep{\fill}}lccccccccc@{}}
\hline
& \multicolumn{3}{c}{\textbf{\# Headlines}} & \multicolumn{3}{c}{\textbf{\# Energy
headlines}} & \multicolumn{3}{c@{}}{\textbf{\% Energy headlines}} \\[-6pt]
& \multicolumn{3}{c}{\hrulefill} & \multicolumn{3}{c}{\hrulefill} &
\multicolumn{3}{c@{}}{\hrulefill} \\
\textbf{Year} & \textbf{NYT} & \textbf{WSJ} & \textbf{Total} &
\textbf{NYT} & \textbf{WSJ} & \textbf{Total} & \textbf{NYT} & \textbf{WSJ} & \textbf{Total} \\
\hline
2008 & \phantom{0}58,951 & 70,905 & 129,856 & 555 & 1869 & 2424 & 0.9 & 2.6 &
1.9 \\
2009 & \phantom{0}47,817 & 78,538 & 126,355 & 287 & 1670 & 1957 & 0.6 & 2.1 &
1.5 \\
2010 & \phantom{0}69,680 & 61,122 & 130,802 & 661 & 1451 & 2112 & 0.9 & 2.4 &
1.6 \\
2011 & 112,595 & 73,417 & 186,012 & 959 & 1615 & 2574 & 0.9 & 2.2 &
1.4 \\[3pt]
All & 289,293 & 284,031 & 573,324 & 2462 & 6605 & 9067 & 0.9 & 2.3
& 1.6 \\
\hline
\end{tabular*}
\end{table}

To effectively remove differences in style, we can select different
baselines for comparison.
In particular, we conduct a ``difference of differences'' approach by
(1) comparing NYT energy headlines to NYT nonenergy headlines to
``subtract out'' general trends in NYT style, (2) doing the same for
the WSJ, and (3) comparing the two resulting summaries to each other.
In particular, to do this second-phase ``within-source'' analysis, we,
within the NYT headlines only, labeled energy-related headlines as
$+1$, left the rest as baseline ($-1$), and applied CCS. We then did
the same for the WSJ.

This gives two summaries for each year, and two for the overall
comparison. We then directly read and compared these lists. We see some
of the same words in the resulting lists as our head-to-head analysis,
but generally have other, more content-specific words that give a
richer picture.
Note, for the NYT, ``renewable,'' ``greenhouse,'' ``shale'' and
``pipeline.'' The style-based words do not tend to appear.
The within-WSJ comparison produces an overlapping set of words to the
NYT comparison, indicating similar coverage between the two sources:
see ``renewable'' there as well. The differences are, however,
suggestive: ``greenhouse'' is indicated for the NYT each year, and the
WSJ in 2009 only. OPEC appears in 2008--2010 for the WSJ, and only in
2010 for the NYT.

By shifting what the baseline is (in this case comparing the energy
headlines of the NYT to the nonenergy headlines of the NYT instead of
the energy headlines of the WSJ), different aspects of the topic, and
different aspects of the corpus, are foregrounded. In the
``within-source'' comparison, we come to understand in general what
energy headlines are for the respective sources. In the
``between-source'' comparison we focus specifically on what
differentiates the two outlets, which foregrounds style of writing as
well as differential coverage of content. Looking at both seems
important for beginning to understand how these themes play out in the media.

\section{Conclusions}
\label{sec:conclusion}

News media significantly impacts our day-to-day lives, public knowledge
and the direction of public policy. Analyzing the news, however, is a
complicated task. The labor intensity of hand coding and the amount of
news available strongly motivate automated methods.

We proposed a sparse predictive framework for extracting meaningful
summaries of specific subjects or topics from document corpora.
These summaries are contrast-based, built by comparing two collections
of documents to each other and identifying how a primary set differs
from a baseline set.
This concise and comparative summarization (CCS) framework expands the
horizon of possible approaches to text data mining.
We offer it as an example of a simpler method that is potentially more
manipulable, flexible and interpretable than those based on generative models.
In general, we believe that there is a rich area between similar naive
methods, such as simple counts and more heavyweight methods such as LDA.
Sparse regression, at the heart of CCS, lies in this area and has much
to offer.

To better understand the performance of our approach, and to
appropriately tune it to maximize the quality and usability of the
summaries produced, we conducted a human validation experiment to
evaluate different summarizers based on human understanding.
Based on the human experiment, we conclude that features selected using
a sparse prediction framework can generate informative key-phrase
summaries for subjects of interest.

We also found these summaries to be superior to those from simpler
methods currently in wide use such as Co-occurrence.
In particular, the Lasso is a good overall feature selector, quite
robust to how the data is preprocessed and computationally scalable.
When not using the Lasso, proper data preparation is quite important.
In this case, tf-idf is a good overall choice for article-length
documents, but not when the document units are small (e.g., paragraphs
and, presumably, headlines, online comments and tweets), in which case
an $L^2$ scaling should be used.\looseness=1

We illustrated the use of our summarizers by evaluating two media
framing questions.
The summarizers indeed allowed for insight and evidence collection.
One of the key aspects of our tool is its interactive capacity; a
researcher can easily work with resulting summary phrases, using them
as topics in their own right, adding them to the concept of the
original topic or dropping them altogether.
Overall, we argue that CCS allows researchers to easily explore large
corpora of documents with an eye to obtaining concise portrayals of any
subject they desire.
A shortcoming of the tool is that both generating the labeling and
interpreting resulting phrases can depend on fairly detailed knowledge
of the topic being explored.
To help with this, we are currently extending the tool to allow for
sentence selection so researchers can contextualize the phrases more rapidly.\looseness=1


\section*{Acknowledgments}
We are indebted to the staff of the XLab at UC Berkeley for their help in
planning and conducting the human validation study.
We are also grateful to Hoxie Ackerman and Saheli Datta for
help assembling this publication.
Luke Miratrix is grateful for the support of a
Graduate Research Fellowship from the National Science Foundation.
Jinzhu Jia's work was done when he was a postdoc at UC Berkeley,
supplemented by NSF SES-0835531.


%


\printaddresses


\begin{thebibliography}{45}

\bibitem[\protect\citeauthoryear{Bischof and Airoldi}{2012}]{bischof2012}
%
\begin{bmisc}[author]
\bauthor{\bsnm{Bischof},~\bfnm{Jonathan~M.}\binits{J.~M.}} \AND
\bauthor{\bsnm{Airoldi},~\bfnm{Edoardo~M.}\binits{E.~M.}}
(\byear{2012}).
\bhowpublished{Summarizing topical content with word frequency and exclusivity.
In \textit{Proceedings of the 29th International Conference on Machine Learning (ICML-12)} 201--208. Edinburgh, Scotland.}
\end{bmisc}
%
\bptok{imsref}%
\endbibitem

\bibitem[\protect\citeauthoryear{Blei and McAuliffe}{2008}]{BleiMcA08}
%
\begin{bincollection}[author]
\bauthor{\bsnm{Blei},~\bfnm{David}\binits{D.}} \AND
\bauthor{\bsnm{McAuliffe},~\bfnm{Jon}\binits{J.}}
(\byear{2008}).
\btitle{Supervised topic models}.
In \bbooktitle{Advances in Neural Information Processing Systems 20}
(\beditor{\bfnm{J.~C.}\binits{J.~C.}~\bsnm{Platt}},
\beditor{\bfnm{D.}\binits{D.}~\bsnm{Koller}},
\beditor{\bfnm{Y.}\binits{Y.}~\bsnm{Singer}} \AND
\beditor{\bfnm{S.}\binits{S.}~\bsnm{Roweis}}, eds.)
\bpages{121--128}.
\bpublisher{MIT Press},
\blocation{Cambridge, MA}.
\end{bincollection}
%
\bptok{imsref}%
\endbibitem

\bibitem[\protect\citeauthoryear{Blei, Ng and Jordan}{2003}]{Blei03}
%
\begin{barticle}[author]
\bauthor{\bsnm{Blei},~\bfnm{David~M.}\binits{D.~M.}},
\bauthor{\bsnm{Ng},~\bfnm{Andrew~Y.}\binits{A.~Y.}} \AND
\bauthor{\bsnm{Jordan},~\bfnm{Michael~I.}\binits{M.~I.}}
(\byear{2003}).
\btitle{Latent {D}irichlet allocation}.
\bjournal{J. Mach. Learn. Res.}
\bvolume{3}
\bpages{993--1022}.
\end{barticle}
%
\bptok{imsref}%
\endbibitem

\bibitem[\protect\citeauthoryear{Chang et~al.}{2009}]{chang09}
%
\begin{binproceedings}[author]
\bauthor{\bsnm{Chang},~\bfnm{Jonathan}\binits{J.}},
\bauthor{\bsnm{Boyd-Graber},~\bfnm{Jordan}\binits{J.}},
\bauthor{\bsnm{Gerrish},~\bfnm{Sean}\binits{S.}},
\bauthor{\bsnm{Wang},~\bfnm{Chong}\binits{C.}} \AND
\bauthor{\bsnm{Blei},~\bfnm{David}\binits{D.}}
(\byear{2009}).
\btitle{Reading tea leaves: How humans interpret topic models}.
In \bbooktitle{Advances in Neural Information Processing Systems 22}
(\beditor{\bfnm{Y.}\binits{Y.}~\bsnm{Bengio}},
\beditor{\bfnm{D.}\binits{D.}~\bsnm{Schuurmans}},
\beditor{\bfnm{J.}\binits{J.}~\bsnm{Lafferty}},
\beditor{\bfnm{C.~K.~I.}\binits{C.~K.~I.}~\bsnm{Williams}} \AND
\beditor{\bfnm{A.}\binits{A.}~\bsnm{Culotta}}, eds.)
\bpages{288--296}.
\blocation{Vancouver, BC, Canada}.
\end{binproceedings}
%
\bptok{imsref}%
\endbibitem


\bibitem[\protect\citeauthoryear{Clavier et~al.}{2010}]{chinapaper}
%
\begin{bmisc}[author]
\bauthor{\bsnm{Clavier},~\bfnm{Sophia}\binits{S.}},
\bauthor{\bsnm{El~Ghaoui},~\bfnm{Laurent}\binits{L.}},
\bauthor{\bsnm{Barnesmoore},~\bfnm{Luke}\binits{L.}} \AND
\bauthor{\bsnm{Li},~\bfnm{Guan-Cheng}\binits{G.-C.}}
(\byear{2010}).
\bhowpublished{All the news that's fit to compare: Comparing Chinese
representations in the American Press and US representations in the
Chinese press}.
\end{bmisc}
%
\bptok{imsref}%
\endbibitem

\bibitem[\protect\citeauthoryear{Dai et~al.}{2011}]{xinyuetal11}
%
\begin{binproceedings}[author]
\bauthor{\bsnm{Dai},~\bfnm{Xinyu}\binits{X.}},
\bauthor{\bsnm{Jia},~\bfnm{Jinzhu}\binits{J.}},
\bauthor{\bsnm{{El Ghaoui}},~\bfnm{Laurent}\binits{L.}} \AND
\bauthor{\bsnm{Yu.},~\bfnm{Bin}\binits{B.}}
(\byear{2011}).
\btitle{{SBA}-term: Sparse bilingual association for terms}.
In \bbooktitle{Fifth IEEE International Conference on Semantic Computing (ICSC)}
\bpages{189--192}.
\bpublisher{Stanford Univ.},
\blocation{Palo Alto, CA}.
\end{binproceedings}
%
\bptok{imsref}%
\endbibitem

\bibitem[\protect\citeauthoryear{Eisenstein, Smith and
Xing}{2011}]{eisenstein2011discovering}
%
\begin{binproceedings}[author]
\bauthor{\bsnm{Eisenstein},~\bfnm{Jacob}\binits{J.}},
\bauthor{\bsnm{Smith},~\bfnm{Noah~A.}\binits{N.~A.}} \AND
\bauthor{\bsnm{Xing},~\bfnm{Eric~P.}\binits{E.~P.}}
(\byear{2011}).
\btitle{Discovering sociolinguistic associations with structured sparsity}.
In \bbooktitle{Proceedings of the 49th Annual Meeting of the
Association for Computational Linguistics: Human Language
Technologies}
\bvolume{1}
\bpages{1365--1374}.
\bpublisher{Association for Computational Linguistics},
\blocation{Portland, OR}.
\end{binproceedings}
%
\bptok{imsref}%
\endbibitem

\bibitem[\protect\citeauthoryear{{El Ghaoui}, Viallon and
Rabbani}{2010}]{elghaoui11}
%
\begin{bmisc}[author]
\bauthor{\bsnm{{El Ghaoui}},~\bfnm{Laurent}\binits{L.}},
\bauthor{\bsnm{Viallon},~\bfnm{Vivian}\binits{V.}} \AND
\bauthor{\bsnm{Rabbani},~\bfnm{Tarek}\binits{T.}}
(\byear{2010}).
\bhowpublished{Safe feature elimination in sparse supervised learning.
Technical Report No. UC/EECS-2010-126.
EECS Dept., Univ. California, Berkeley}.
\end{bmisc}
%
\bptok{imsref}%
\endbibitem

\bibitem[\protect\citeauthoryear{{El Ghaoui} et~al.}{2011}]{ELDPSB:11}
%
\begin{binproceedings}[author]
\bauthor{\bsnm{{El Ghaoui}},~\bfnm{Laurent}\binits{L.}},
\bauthor{\bsnm{Li},~\bfnm{Guan-Cheng}\binits{G.-C.}},
\bauthor{\bsnm{Duong},~\bfnm{Viet-An}\binits{V.-A.}},
\bauthor{\bsnm{Pham},~\bfnm{Vu}\binits{V.}},
\bauthor{\bsnm{Srivastava},~\bfnm{Ashok}\binits{A.}} \AND
\bauthor{\bsnm{Bhaduri},~\bfnm{Kanishka}\binits{K.}}
(\byear{2011}).
\btitle{Sparse machine learning methods for understanding large text
corpora: Application to flight reports}.
In \bbooktitle{Conference on Intelligent Data Understanding}
\bpages{159--173}.
\blocation{Mountain View, CA}.
\end{binproceedings}
%
\bptok{imsref}%
\endbibitem

\bibitem[\protect\citeauthoryear{Entman}{1993}]{entman1993}
%
\begin{barticle}[author]
\bauthor{\bsnm{Entman},~\bfnm{R.~M.}\binits{R.~M.}}
(\byear{1993}).
\btitle{Framing: Toward clarification of a fractured paradigm}.
\bjournal{Journal of Communication}
\bvolume{43}
\bpages{52--57}.
\end{barticle}
%
\bptok{imsref}%
\endbibitem

\bibitem[\protect\citeauthoryear{Entman}{2004}]{entman2004}
%
\begin{bbook}[author]
\bauthor{\bsnm{Entman},~\bfnm{R.~M.}\binits{R.~M.}}
(\byear{2004}).
\btitle{Projections of power framing news, public opinion, and U.S.
foreign policy}.
\bpublisher{Univ. Chicago},
\blocation{Chicago, IL}.
\end{bbook}
%
\bptok{imsref}%
\endbibitem

\bibitem[\protect\citeauthoryear{Forman}{2003}]{foreman2003bns}
%
\begin{barticle}[author]
\bauthor{\bsnm{Forman},~\bfnm{George}\binits{G.}}
(\byear{2003}).
\btitle{An extensive empirical study of feature selection metrics for
text classification}.
\bjournal{J. Mach. Learn. Res.}
\bvolume{3}
\bpages{1289--1305}.
\end{barticle}
%
\bptok{imsref}%
\endbibitem

\bibitem[\protect\citeauthoryear{Frank et~al.}{1999}]{franketal1999}
%
\begin{binproceedings}[author]
\bauthor{\bsnm{Frank},~\bfnm{Eibe}\binits{E.}},
\bauthor{\bsnm{Paynter},~\bfnm{Gordon~W.}\binits{G.~W.}},
\bauthor{\bsnm{Witten},~\bfnm{Ian~H.}\binits{I.~H.}},
\bauthor{\bsnm{Gutwin},~\bfnm{Carl}\binits{C.}} \AND
\bauthor{\bsnm{Nevill-Manning},~\bfnm{Craig~G.}\binits{C.~G.}}
(\byear{1999}).
\btitle{Domain-specific keyphrase extraction}.
In \bbooktitle{Proceedings of the Sixteenth International Joint
Conference on Artificial Intelligence (IJCAI-99)}
\bpages{668--673}.
\bpublisher{Morgan Kaufmann},
\blocation{San Francisco, CA}.
\end{binproceedings}
%
\bptok{imsref}%
\endbibitem

\bibitem[\protect\citeauthoryear{Gawalt et~al.}{2010}]{mir}
%
\begin{binproceedings}[author]
\bauthor{\bsnm{Gawalt},~\bfnm{Brian}\binits{B.}},
\bauthor{\bsnm{Jia},~\bfnm{Jinzhu}\binits{J.}},
\bauthor{\bsnm{Miratrix},~\bfnm{Luke~W.}\binits{L.~W.}},
\bauthor{\bsnm{Ghaoui},~\bfnm{Laurent}\binits{L.}},
\bauthor{\bsnm{Yu},~\bfnm{Bin}\binits{B.}} \AND
\bauthor{\bsnm{Clavier},~\bfnm{Sophia}\binits{S.}}
(\byear{2010}).
\btitle{Discovering word associations in news media via feature
selection and sparse classi{fi}cation}.
In \bbooktitle{Proceedings of the International Conference on Multimedia
Information Retrieval (MIR'10)}
\bpages{211--220}.
\blocation{Philadelphia, PA}.
\end{binproceedings}
%
\bptok{imsref}%
\endbibitem

\bibitem[\protect\citeauthoryear{Genkin, Lewis and Madigan}{2007}]{genkin07}
%
\begin{barticle}[mr]
\bauthor{\bsnm{Genkin},~\bfnm{Alexander}\binits{A.}},
\bauthor{\bsnm{Lewis},~\bfnm{David~D.}\binits{D.~D.}} \AND
\bauthor{\bsnm{Madigan},~\bfnm{David}\binits{D.}}
(\byear{2007}).
\btitle{Large-scale {B}ayesian logistic regression for text categorization}.
\bjournal{Technometrics}
\bvolume{49}
\bpages{291--304}.
\bid{doi={10.1198/004017007000000245}, issn={0040-1706}, mr={2408634}}
\end{barticle}
%
\bptok{imsref}%
\endbibitem

\bibitem[\protect\citeauthoryear{Goffman}{1974}]{goffman1974}
%
\begin{bbook}[author]
\bauthor{\bsnm{Goffman},~\bfnm{E.}\binits{E.}}
(\byear{1974}).
\btitle{Frame Analysis: An Essay on the Organization of Experience}.
\bpublisher{Harvard Univ. Press},
\blocation{Cambridge, MA}.
\end{bbook}
%
\bptok{imsref}%
\endbibitem

\bibitem[\protect\citeauthoryear{Goldstein et~al.}{2000}]{goldstein2000}
%
\begin{binproceedings}[author]
\bauthor{\bsnm{Goldstein},~\bfnm{Jade}\binits{J.}},
\bauthor{\bsnm{Mittal},~\bfnm{Vibhu}\binits{V.}},
\bauthor{\bsnm{Carbonell},~\bfnm{Jaime}\binits{J.}} \AND
\bauthor{\bsnm{Kantrowitz},~\bfnm{Mark}\binits{M.}}
(\byear{2000}).
\btitle{Multi-document summarization by sentence extraction}.
In \bbooktitle{NAACL-ANLP 2000 Workshop on Automatic Summarization}
\bpages{40--48}.
\blocation{Seattle, WA}.
\end{binproceedings}
%
\bptok{imsref}%
\endbibitem

\bibitem[\protect\citeauthoryear{Grimmer et~al.}{2011}]{grimmeretal2011}
%
\begin{bmisc}[author]
\bauthor{\bsnm{Grimmer},~\bfnm{Justin}\binits{J.}},
\bauthor{\bsnm{Shorey},~\bfnm{Rachel}\binits{R.}},
\bauthor{\bsnm{Wallach},~\bfnm{Hanna}\binits{H.}} \AND
\bauthor{\bsnm{Zlotnick},~\bfnm{Frances}\binits{F.}}
(\byear{2011}).
\bhowpublished{A class of Bayesian semiparametric cluster-topic models for
political texts}.
\end{bmisc}
%
\bptok{imsref}%
\endbibitem

\bibitem[\protect\citeauthoryear{Hastie, Tibshirani and Friedman}{2011}]{HTF03}
%
\begin{bbook}[author]
\bauthor{\bsnm{Hastie},~\bfnm{T.}\binits{T.}},
\bauthor{\bsnm{Tibshirani},~\bfnm{R.}\binits{R.}} \AND
\bauthor{\bsnm{Friedman},~\bfnm{J.~H.}\binits{J.~H.}}
(\byear{2011}).
\btitle{The Elements of Statistical Learning, Vol. 1}.
\bpublisher{Springer},
\blocation{New York}.
\end{bbook}
%
\bptok{imsref}%
\endbibitem

\bibitem[\protect\citeauthoryear{Hennig}{2009}]{hennig2009}
%
\begin{binproceedings}[author]
\bauthor{\bsnm{Hennig},~\bfnm{Leonhard}\binits{L.}}
(\byear{2009}).
\btitle{Topic-based multi-document summarization with probabilistic
latent semantic analysis}.
In \bbooktitle{Recent Advances in Natural Language Processing (RANLP)}
\bpages{144--149}.
\bpublisher{Association for Computational Linguistics}, \blocation{Borovets, Bulgaria}.
\end{binproceedings}
%
\bptok{imsref}%
\endbibitem

\bibitem[\protect\citeauthoryear{Hopkins and King}{2010}]{HopKin10}
%
\begin{barticle}[author]
\bauthor{\bsnm{Hopkins},~\bfnm{Daniel}\binits{D.}} \AND
\bauthor{\bsnm{King},~\bfnm{Gary}\binits{G.}}
(\byear{2010}).
\btitle{A method of automated nonparametric content analysis for social
science}.
\bjournal{American Journal of Political Science}
\bvolume{54}
\bpages{229--247}.
\end{barticle}
%
\bptok{imsref}%
\endbibitem

\bibitem[\protect\citeauthoryear{Ifrim, Bakir and Weikum}{2008}]{ifrim08}
%
\begin{binproceedings}[author]
\bauthor{\bsnm{Ifrim},~\bfnm{Georgiana}\binits{G.}},
\bauthor{\bsnm{Bakir},~\bfnm{G{\"{o}}khan}\binits{G.}} \AND
\bauthor{\bsnm{Weikum},~\bfnm{Gerhard}\binits{G.}}
(\byear{2008}).
\btitle{Fast logistic regression for text categorization with
variable-length N-grams}.
In \bbooktitle{14th ACM SIGKDD International Conference on Knowledge
Discovery and Data Mining}
\bpages{354--362}.
\bpublisher{ACM},
\blocation{New York}.
\end{binproceedings}
%
\bptok{imsref}%
\endbibitem


\bibitem[\protect\citeauthoryear{Jia et~al.}{2011}]{techreport}
%
\begin{bmisc}[author]
\bauthor{\bsnm{Jia},~\bfnm{Jinzhu}\binits{J.}},
\bauthor{\bsnm{Miratrix},~\bfnm{Luke~W.}\binits{L.~W.}},
\bauthor{\bsnm{Gawalt},~\bfnm{Brian}\binits{B.}},
\bauthor{\bsnm{Yu},~\bfnm{Bin}\binits{B.}} \AND
\bauthor{\bsnm{El~Ghaoui},~\bfnm{Laurent}\binits{L.}}
(\byear{2011}).
\bhowpublished{What is in the news on a subject: Automatic and sparse
summarization of large document corpora.
Technical Report \#801, Dept. Statistics, Univ. California, Berkeley}.
\end{bmisc}
%
\bptok{imsref}%
\endbibitem\

\bibitem[\protect\citeauthoryear{Kiousis and Wu}{2008}]{kiousisetal2008}
%
\begin{barticle}[author]
\bauthor{\bsnm{Kiousis},~\bfnm{Spiro}\binits{S.}} \AND
\bauthor{\bsnm{Wu},~\bfnm{Xu}\binits{X.}}
(\byear{2008}).
\btitle{International agenda-building and agenda-setting: Exploring the
influence of public relations counsel on US news media and public
perceptions of foreign nations}.
\bjournal{The International Communications Gazette}
\bvolume{70}
\bpages{58--75}.
\end{barticle}
%
\bptok{imsref}%
\endbibitem

\bibitem[\protect\citeauthoryear{Kunczik}{2000}]{kunczik2000}
%
\begin{bmisc}[author]
\bauthor{\bsnm{Kunczik},~\bfnm{Michael}\binits{M.}}
(\byear{2000}).
\bhowpublished{Globalisation: News media, images of nations and the flow of
international capital with special reference to the role of rating agencies.
\textit{J. International Communication} \textbf{8} 39--79}.
\end{bmisc}
%
\bptok{imsref}%
\endbibitem

\bibitem[\protect\citeauthoryear{Lazer et~al.}{2009}]{lazeretal09}
%
\begin{barticle}[pbm]
\bauthor{\bsnm{Lazer},~\bfnm{David}\binits{D.}},
\bauthor{\bsnm{Pentland},~\bfnm{Alex}\binits{A.}},
\bauthor{\bsnm{Adamic},~\bfnm{Lada}\binits{L.}},
\bauthor{\bsnm{Aral},~\bfnm{Sinan}\binits{S.}},
\bauthor{\bsnm{Barabasi},~\bfnm{Albert-Laszlo}\binits{A.-L.}},
\bauthor{\bsnm{Brewer},~\bfnm{Devon}\binits{D.}},
\bauthor{\bsnm{Christakis},~\bfnm{Nicholas}\binits{N.}},
\bauthor{\bsnm{Contractor},~\bfnm{Noshir}\binits{N.}},
\bauthor{\bsnm{Fowler},~\bfnm{James}\binits{J.}},
\bauthor{\bsnm{Gutmann},~\bfnm{Myron}\binits{M.}},
\bauthor{\bsnm{Jebara},~\bfnm{Tony}\binits{T.}},
\bauthor{\bsnm{King},~\bfnm{Gary}\binits{G.}},
\bauthor{\bsnm{Macy},~\bfnm{Michael}\binits{M.}},
\bauthor{\bsnm{Roy},~\bfnm{Deb}\binits{D.}} \AND
\bauthor{\bparticle{Van} \bsnm{Alstyne},~\bfnm{Marshall}\binits{M.}}
(\byear{2009}).
\btitle{Computational social science}.
\bjournal{Science}
\bvolume{323}
\bpages{721--723}.
\bid{doi={10.1126/science.1167742}, issn={1095-9203}, mid={NIHMS98137},
pii={323/5915/721}, pmcid={2745217}, pmid={19197046}}
\end{barticle}
%
\bptok{imsref}%
\endbibitem

\bibitem[\protect\citeauthoryear{Lee and Chen}{2006}]{lee2006nmt}
%
\begin{barticle}[author]
\bauthor{\bsnm{Lee},~\bfnm{L.}\binits{L.}} \AND
\bauthor{\bsnm{Chen},~\bfnm{S.}\binits{S.}}
(\byear{2006}).
\btitle{New methods for text categorization based on a new feature
selection method and a new similarity measure between documents}.
\bjournal{Lecture Notes in Comput. Sci.}
\bvolume{4031}
\bpages{1280}.
\end{barticle}
%
\bptok{imsref}%
\endbibitem

\bibitem[\protect\citeauthoryear{McLeod, Kosicki and Pan}{1991}]{mcleodetal1991}
%
\begin{bbook}[author]
\bauthor{\bsnm{McLeod},~\bfnm{M.}\binits{M.}},
\bauthor{\bsnm{Kosicki},~\bfnm{G.~M.}\binits{G.~M.}} \AND
\bauthor{\bsnm{Pan},~\bfnm{Z.}\binits{Z.}}
(\byear{1991}).
\btitle{On Understanding and Misunderstanding Media Effects}.
\bpublisher{Edward Arnold},
\blocation{London}.
\end{bbook}
%
\bptok{imsref}%
\endbibitem


\bibitem[\protect\citeauthoryear{Monroe, Colaresi and Quinn}{2008}]{monroe08}
%
\begin{barticle}[author]
\bauthor{\bsnm{Monroe},~\bfnm{Burt~L.}\binits{B.~L.}},
\bauthor{\bsnm{Colaresi},~\bfnm{Michael~P.}\binits{M.~P.}} \AND
\bauthor{\bsnm{Quinn},~\bfnm{Kevin~M.}\binits{K.~M.}}
(\byear{2008}).
\btitle{Fightin' words: Lexical feature selection and evaluation for
identifying the content of political conflict}.
\bjournal{Political Analysis}
\bvolume{16}
\bpages{372--403}.
\end{barticle}
%
\bptok{imsref}%
\endbibitem

\bibitem[\protect\citeauthoryear{Mosteller and Wallace}{1984}]{mosteller84}
%
\begin{bbook}[mr]
\bauthor{\bsnm{Mosteller},~\bfnm{Frederick}\binits{F.}} \AND
\bauthor{\bsnm{Wallace},~\bfnm{David~L.}\binits{D.~L.}}
(\byear{1984}).
\btitle{Applied {B}ayesian and Classical Inference:
The Case of The Federalist Papers},
\bedition{2nd}~ed.
\bpublisher{Springer},
\blocation{New York}.
\bid{doi={10.1007/978-1-4612-5256-6}, mr={0766742}}
\end{bbook}
%
\bptok{imsref}%
\endbibitem

\bibitem[\protect\citeauthoryear{Neto, Freitas and
Kaestner}{2002}]{netoetal2002}
%
\begin{bincollection}[mr]
\bauthor{\bsnm{Neto},~\bfnm{Joel~Larocca}\binits{J.~L.}},
\bauthor{\bsnm{Freitas},~\bfnm{Alex~A.}\binits{A.~A.}} \AND
\bauthor{\bsnm{Kaestner},~\bfnm{Celso~A.~A.}\binits{C.~A.~A.}}
(\byear{2002}).
\btitle{Automatic text summarization using a machine learning approach}.
In \bbooktitle{Advances in Artificial Intelligence}.
\bseries{Lecture Notes in Computer Science}
\bvolume{2507}
\bpages{205--215}.
\bpublisher{Springer},
\blocation{Berlin}.
\bid{doi={10.1007/3-540-36127-8_20}, mr={2048852}}
\end{bincollection}
%
\bptok{imsref}%
\endbibitem

\bibitem[\protect\citeauthoryear{Paul, Zhai and
Girju}{2010}]{paul2010summarizing}
%
\begin{binproceedings}[author]
\bauthor{\bsnm{Paul},~\bfnm{Michael~J.}\binits{M.~J.}},
\bauthor{\bsnm{Zhai},~\bfnm{ChengXiang}\binits{C.}} \AND
\bauthor{\bsnm{Girju},~\bfnm{Roxana}\binits{R.}}
(\byear{2010}).
\btitle{Summarizing contrastive viewpoints in opinionated text}.
In \bbooktitle{Proceedings of the 2010 Conference on Empirical Methods
in Natural Language Processing}
\bpages{66--76}.
\bpublisher{Association for Computational Linguistics, Stroudsburg, PA}.
\end{binproceedings}
%
\bptok{imsref}%
\endbibitem

\bibitem[\protect\citeauthoryear{Pottker}{2003}]{Pottker03}
%
\begin{barticle}[author]
\bauthor{\bsnm{Pottker},~\bfnm{H.}\binits{H.}}
(\byear{2003}).
\btitle{News and its communicative quality: The inverted pyramid---when
and why did it appear?}
\bjournal{Journalism Studies}
\bvolume{4}
\bpages{501--511}.
\end{barticle}
%
\bptok{imsref}%
\endbibitem

\bibitem[\protect\citeauthoryear{Rose et~al.}{2010}]{roseetal2010}
%
\begin{bincollection}[author]
\bauthor{\bsnm{Rose},~\bfnm{Stuart}\binits{S.}},
\bauthor{\bsnm{Engel},~\bfnm{Dave}\binits{D.}},
\bauthor{\bsnm{Cramer},~\bfnm{Nick}\binits{N.}} \AND
\bauthor{\bsnm{Cowley},~\bfnm{Wendy}\binits{W.}}
(\byear{2010}).
\btitle{Automatic keyword extraction from individual documents}.
In \bbooktitle{Text Mining: Applications and Theory}
(\beditor{\bfnm{Michael~W.}\binits{M.~W.}~\bsnm{Berry}} \AND
\beditor{\bfnm{Jacob}\binits{J.}~\bsnm{Kogan}}, eds.).
\bpublisher{Wiley},
\blocation{Chichester}.
\end{bincollection}
%
\bptok{imsref}%
\endbibitem

\bibitem[\protect\citeauthoryear{Salton}{1991}]{salton1991dat}
%
\begin{barticle}[pbm]
\bauthor{\bsnm{Salton},~\bfnm{G.}\binits{G.}}
(\byear{1991}).
\btitle{Developments in automatic text retrieval}.
\bjournal{Science}
\bvolume{253}
\bpages{974--980}.
\bid{doi={10.1126/science.253.5023.974}, issn={0036-8075},
pii={253/5023/974}, pmid={17775340}}
\end{barticle}
%
\bptok{imsref}%
\endbibitem

\bibitem[\protect\citeauthoryear{Salton and Buckley}{1988}]{salton88}
%
\begin{barticle}[author]
\bauthor{\bsnm{Salton},~\bfnm{G.}\binits{G.}} \AND
\bauthor{\bsnm{Buckley},~\bfnm{C.}\binits{C.}}
(\byear{1988}).
\btitle{Term-weighting approaches in automatic text retrieval}.
\bjournal{Information Processing and Management}
\bvolume{24}
\bpages{513--523}.
\end{barticle}
%
\bptok{imsref}%
\endbibitem

\bibitem[\protect\citeauthoryear{Senellart and Blondel}{2008}]{senellart2008}
%
\begin{bincollection}[author]
\bauthor{\bsnm{Senellart},~\bfnm{P.}\binits{P.}} \AND
\bauthor{\bsnm{Blondel},~\bfnm{V.~D.}\binits{V.~D.}}
(\byear{2008}).
\btitle{Automatic discovery of similar words}.
In \bbooktitle{Survey of Text Mining II}.
\bpublisher{Springer},
\blocation{Berlin}.
\end{bincollection}
%
\bptok{imsref}%
\endbibitem

\bibitem[\protect\citeauthoryear{Shahaf, Guestrin and
Horvitz}{2012}]{shahaf2012trains}
%
\begin{binproceedings}[author]
\bauthor{\bsnm{Shahaf},~\bfnm{Dafna}\binits{D.}},
\bauthor{\bsnm{Guestrin},~\bfnm{Carlos}\binits{C.}} \AND
\bauthor{\bsnm{Horvitz},~\bfnm{Eric}\binits{E.}}
(\byear{2012}).
\btitle{Trains of thought: Generating information maps}.
In \bbooktitle{Proceedings of the 21st International Conference on
World Wide Web}
\bpages{899--908}.
\bpublisher{ACM},
\blocation{Lyon, France}.
\end{binproceedings}
%
\bptok{imsref}%
\endbibitem

\bibitem[\protect\citeauthoryear{Tibshirani}{1996}]{tibshirani1996regression}
%
\begin{barticle}[mr]
\bauthor{\bsnm{Tibshirani},~\bfnm{Robert}\binits{R.}}
(\byear{1996}).
\btitle{Regression shrinkage and selection via the lasso}.
\bjournal{J. R. Stat. Soc. Ser. B Stat. Methodol.}
\bvolume{58}
\bpages{267--288}.
\bid{issn={0035-9246}, mr={1379242}}
\end{barticle}
%
\bptok{imsref}%
\endbibitem

\bibitem[\protect\citeauthoryear{Wagstaff}{2012}]{Wagstaff:2012tr}
%
\begin{binproceedings}[author]
\bauthor{\bsnm{Wagstaff},~\bfnm{Kiri~L.}\binits{K.~L.}}
(\byear{2012}).
\btitle{{Machine learning that matters}}.
In \bbooktitle{29th International Conference on Machine Learning}
\bpages{1--6}.
\blocation{Edinburgh, Scotland}.
\end{binproceedings}
%
\bptok{imsref}%
\endbibitem

\bibitem[\protect\citeauthoryear{Yang and Pendersen}{1997}]{yang1997csf}
%
\begin{binproceedings}[author]
\bauthor{\bsnm{Yang},~\bfnm{Y.}\binits{Y.}} \AND
\bauthor{\bsnm{Pendersen},~\bfnm{I.~O.}\binits{I.~O.}}
(\byear{1997}).
\btitle{A comparative study on feature selection in text categorization}.
In \bbooktitle{ICML-97, 14th International Conference on Machine Learning}
\bpages{412--420}.
\blocation{Nashville, TN}.
\end{binproceedings}
%
\bptok{imsref}%
\endbibitem

\bibitem[\protect\citeauthoryear{Zhang and Oles}{2001}]{zhang01}
%
\begin{barticle}[author]
\bauthor{\bsnm{Zhang},~\bfnm{Tong}\binits{T.}} \AND
\bauthor{\bsnm{Oles},~\bfnm{Frank~J.}\binits{F.~J.}}
(\byear{2001}).
\btitle{Text categorization based on regularized linear classfiication methods}.
\bjournal{Information Retrieval}
\bvolume{4}
\bpages{5--31}.
\end{barticle}
%
\bptok{imsref}%
\endbibitem

\bibitem[\protect\citeauthoryear{Zhao and Yu}{2007}]{zhao2007stagewise}
%
\begin{barticle}[mr]
\bauthor{\bsnm{Zhao},~\bfnm{Peng}\binits{P.}} \AND
\bauthor{\bsnm{Yu},~\bfnm{Bin}\binits{B.}}
(\byear{2007}).
\btitle{Stagewise lasso}.
\bjournal{J. Mach. Learn. Res.}
\bvolume{8}
\bpages{2701--2726}.
\bid{issn={1532-4435}, mr={2383572}}
\end{barticle}
%
\bptok{imsref}%
\endbibitem\

\bibitem[\protect\citeauthoryear{Zubiaga et~al.}{2011}]{Zubiaga11}
%
\begin{binproceedings}[author]
\bauthor{\bsnm{Zubiaga},~\bfnm{Arkaitz}\binits{A.}},
\bauthor{\bsnm{Spina},~\bfnm{Damiano}\binits{D.}},
\bauthor{\bsnm{Fresno},~\bfnm{V{\'{\i}}ctor}\binits{V.}} \AND
\bauthor{\bsnm{Mart{\'{\i}}nez},~\bfnm{Raquel}\binits{R.}}
(\byear{2011}).
\btitle{Classifying trending topics: A typology of conversation
triggers on Twitter}.
In \bbooktitle{Proceedings of the 20th ACM International Conference on
Information and Knowledge Management (CIKM'11)}
\bpages{2461--2464}.
\bpublisher{ACM},
\blocation{New York}.
\end{binproceedings}
%
\bptok{imsref}%
\endbibitem

\end{thebibliography}
\end{document}